\documentclass[11pt,a4paper]{article}

\usepackage[T1]{fontenc}
\usepackage[utf8]{inputenc}
\usepackage[english]{babel}
\usepackage{helvet}                 

\usepackage[a4paper,margin=2.5cm]{geometry}
\usepackage{setspace}
\usepackage{microtype}
\usepackage{graphicx}
\usepackage{booktabs}
\usepackage{array}
\usepackage{caption}
\usepackage{pdflscape}
\usepackage{lscape}
\usepackage[modulo]{lineno}
\usepackage{enumitem}
\usepackage{xcolor}
\usepackage[numbers,super,sort&compress]{natbib}
\usepackage[colorlinks=true,allcolors=blue,breaklinks=true]{hyperref}
\usepackage{xurl}

\setlist[itemize]{leftmargin=1.6em,itemsep=2pt,topsep=2pt}

\newcommand{\mdot}{\textperiodcentered}

\usepackage{titlesec}
\definecolor{lancetred}{RGB}{192,0,0}
\titleformat{\section}{\normalfont\large\bfseries\color{lancetred}}{}{0pt}{}
\titleformat{\subsection}{\normalfont\normalsize\bfseries}{}{0pt}{}
\titlespacing*{\section}{0pt}{14pt}{6pt}
\titlespacing*{\subsection}{0pt}{10pt}{4pt}

\hypersetup{
  pdftitle={Clinician-Friendly Foundation Models for Ophthalmic Image Diagnostics without Fine-Tuning or Technical Barriers},
  pdfauthor={Meng Wang et al.}
}

\begin{document}

\onehalfspacing

\begin{center}
{\LARGE\bfseries Clinician-Friendly Foundation Models for Ophthalmic Image Diagnostics without Fine-Tuning or Technical Barriers\par}

\vspace{1.2em}

{\normalsize
Meng Wang\textsuperscript{1,2,\&},
Tian Lin\textsuperscript{3,4,\&},
Qingshan Hou\textsuperscript{1,2,5,\&},
Aidi Lin\textsuperscript{3,6,\&},
Lianyu Wang\textsuperscript{1,2,\&},
Jingcheng Wang\textsuperscript{7},
Qingsheng Peng\textsuperscript{8},
Truong X. Nguyen\textsuperscript{9,10},
Zhi Da Soh\textsuperscript{8},
Xiayin Zhang\textsuperscript{8},
Jingyan Yang\textsuperscript{11},
Danqi Fang\textsuperscript{9},
Ke Zou\textsuperscript{1,2},
Ting Xu\textsuperscript{1,2},
Can Can Xue\textsuperscript{8},
Ten Cheer Quek\textsuperscript{8},
Qinkai Yu\textsuperscript{12},
Minxin Liu\textsuperscript{1,2},
Hui Zhou\textsuperscript{13},
Zixuan Xiao\textsuperscript{14},
Guiqin He\textsuperscript{3,6,15},
Huiyu Liang\textsuperscript{16},
Tingkun Shi\textsuperscript{3,6},
Man Chen\textsuperscript{3,6},
Zhuangling Lin\textsuperscript{3},
Linna Liu\textsuperscript{17},
Yuanyuan Peng\textsuperscript{18},
Li Jia Chen\textsuperscript{9},
Chi Ming Chan\textsuperscript{19,20},
Xiaohong Li\textsuperscript{21},
Junren He\textsuperscript{22},
Zhirong Xu\textsuperscript{4},
Tingbing Fang\textsuperscript{4},
Yanli Wang\textsuperscript{4},
Qingzhi Wang\textsuperscript{4},
Wenyi Hu\textsuperscript{23,24},
Yujie Wang\textsuperscript{23,24},
Li Li\textsuperscript{23,24,25,26},
Jiaying Ye\textsuperscript{23,24},
Tonghui Ye\textsuperscript{23,24},
Liang Lyu\textsuperscript{23,24},
Yongjian Lu\textsuperscript{27},
Ruoshi Cai\textsuperscript{28},
Yiwen Tang\textsuperscript{29},
Qiuming Hu\textsuperscript{30},
Junhong Chen\textsuperscript{31},
Zhenhua Zhang\textsuperscript{32},
Cheng Chen\textsuperscript{33},
Yitian Zhao\textsuperscript{34},
Dianbo Liu\textsuperscript{1,2},
Jianhua Wu\textsuperscript{16},
Xinjian Chen\textsuperscript{35},
Changqing Zhang\textsuperscript{36},
Xiaojun Wu\textsuperscript{4,37},
Triet Thanh Nguyen\textsuperscript{10},
Yanda Meng\textsuperscript{11,38},
Yalin Zheng\textsuperscript{38,39},
Daoqiang Zhang\textsuperscript{40},
Xiaochun Cao\textsuperscript{41},
Yih Chung Tham\textsuperscript{1,2,8,42},
Ye Zhang\textsuperscript{11},
Ying Han\textsuperscript{43,44,45},
Alvin L Young\textsuperscript{9,46},
Mary Ho\textsuperscript{9,46},
Carmen K M Chan\textsuperscript{9,47},
Clement C Tham\textsuperscript{9},
Zhuoting Zhu\textsuperscript{23,24},
Carol Y. Cheung\textsuperscript{9},
Tien Yin Wong\textsuperscript{8,48},
Huazhu Fu\textsuperscript{49,*},
Haoyu Chen\textsuperscript{3,6,9,*},
Ching-Yu Cheng\textsuperscript{1,2,8,42,*}
\par}
\end{center}

\vspace{0.6em}

{\small\setlength{\parskip}{2pt}\setlength{\parindent}{0pt}
\textsuperscript{1} Centre for Innovation and Precision Eye Health, Yong Loo Lin School of Medicine, National University of Singapore, Singapore 119228, Singapore.

\textsuperscript{2} Department of Ophthalmology, Yong Loo Lin School of Medicine, National University of Singapore, Singapore 119228, Singapore.

\textsuperscript{3} Joint Shantou International Eye Center of Shantou University and the Chinese University of Hong Kong, Shantou, Guangdong 515041, China.

\textsuperscript{4} Shenzhen Nanshan People's Hospital, 518052 Shenzhen, Guangdong, China.

\textsuperscript{5} School of Computer Science and Engineering, Northeastern University, 110819 Shenyang, Liaoning, China.

\textsuperscript{6} Fifth Clinical Institute of Shantou University Medical College, Shantou, Guangdong 515041, China.

\textsuperscript{7} Big Vision Medical Technology Ltd., 215011 Suzhou, China.

\textsuperscript{8} Singapore Eye Research Institute, Singapore National Eye Centre, Singapore 169856, Singapore.

\textsuperscript{9} Department of Ophthalmology and Visual Sciences, The Chinese University of Hong Kong, 999077 Hong Kong Special Administrative Region, China.

\textsuperscript{10} Binh Dinh Eye Hospital, Vietnam.

\textsuperscript{11} Beijing Tongren Eye Center, Beijing Tongren Hospital, Capital Medical University, Beijing, China.

\textsuperscript{12} Department of Computer Science, University of Exeter, Exeter, EX4 4RN, UK.

\textsuperscript{13} Southern University of Science and Technology Hospital, 518000 Shenzhen, Guangdong, China.

\textsuperscript{14} Department of Ophthalmology \& Visual Sciences, The Chinese University of Hong Kong, Shenzhen Medical Centre, 518100 Shenzhen, Guangdong, China.

\textsuperscript{15} Department of Ophthalmology, Meizhou People's Hospital, Meizhou Academy of Medical Sciences, 514000 Meizhou, Guangdong, China.

\textsuperscript{16} Department of Ophthalmology, Yeungnam University College of Medicine, 42417 Daegu, South Korea.

\textsuperscript{17} Aier Eye Hospital of Wuhan University, 430000 Wuhan, Hubei, China.

\textsuperscript{18} School of Biomedical Engineering, Anhui Medical University, 230032 Hefei, Anhui, China.

\textsuperscript{19} Department of Ophthalmology, Cardinal Tien Hospital, New Taipei City, Taiwan.

\textsuperscript{20} Department of Ophthalmology, School of Medicine, Fu Jen Catholic University, New Taipei City, Taiwan.

\textsuperscript{21} The University of Hong Kong-Shenzhen Hospital, 518053 Shenzhen, Guangdong, China.

\textsuperscript{22} Ophthalmology Department, The Tenth Affiliated Hospital, Southern Medical University (Dongguan people's hospital), 523000 Dongguan, China.

\textsuperscript{23} Centre for Eye Research Australia, Melbourne, Victoria, Australia.

\textsuperscript{24} Department of Surgery (Ophthalmology), University of Melbourne, Melbourne, Victoria, Australia.

\textsuperscript{25} Department of Ophthalmology, Shengli Clinical Medical College of Fujian Medical University, 350001 Fuzhou, Fujian, China.

\textsuperscript{26} Fuzhou University Affiliated Provincial Hospital, 350001 Fuzhou, Fujian, China.

\textsuperscript{27} First Affiliated Hospital of Shantou University Medical College, Shantou 515041, Guangdong, China.

\textsuperscript{28} Department of Ophthalmology, The Second People's Hospital of Foshan, Foshan 528000, Guangdong, China.

\textsuperscript{29} Department of Ophthalmology, the First Affiliated Hospital of Guilin Medical University, Guilin 541001, Guangxi Zhuang Autonomous Region, China.

\textsuperscript{30} Guangxi Jingliang Eye Hospital, 530021 Nanning, Guangxi, China.

\textsuperscript{31} Puning People's Hospital, 522000 Jieyang, Guangdong, China.

\textsuperscript{32} Qingdao Central Hospital, University of Health and Rehabilitation Sciences (Qingdao Central Hospital), 266042 Qingdao, Shandong, China.

\textsuperscript{33} Department of Electrical and Electronic Engineering, The University of Hong Kong, Hong Kong, China.

\textsuperscript{34} Ningbo Institute of Materials Technology and Engineering, Chinese Academy of Sciences, 315201 Ningbo, Zhejiang, China.

\textsuperscript{35} School of Electronics and Information Engineering, Soochow University, 215006 Suzhou, Jiangsu, China.

\textsuperscript{36} College of Intelligence and Computing, Tianjin University, 300350 Tianjin, China.

\textsuperscript{37} Affiliated Nanshan Hospital of Shenzhen University, 518052 Shenzhen, Guangdong, China.

\textsuperscript{38} Liverpool Centre for Cardiovascular Science, University of Liverpool and Liverpool Heart and Chest Hospital, Liverpool, United Kingdom.

\textsuperscript{39} Department of Eye and Vision Sciences, University of Liverpool, Liverpool, United Kingdom.

\textsuperscript{40} College of Computer Science and Technology, Nanjing University of Aeronautics and Astronautics, 211100 Nanjing, Jiangsu, China.

\textsuperscript{41} The School of Cyber Science and Technology, Shenzhen Campus of Sun Yat-Sen University, Shenzhen, China.

\textsuperscript{42} Ophthalmology \& Visual Sciences Academic Clinical Program (EYE ACP), Duke-NUS Medical School, Singapore 169856, Singapore.

\textsuperscript{43} University of California, San Francisco, San Francisco, CA, United States 94158.

\textsuperscript{44} The Francis I. Proctor Foundation for Research in Ophthalmology.

\textsuperscript{45} Ophthalmology Section, Surgical Service, San Francisco Veterans Affairs Medical Center, San Francisco, California.

\textsuperscript{46} Department of Ophthalmology and Visual Sciences, Prince of Wales Hospital, Shatin, Hong Kong.

\textsuperscript{47} Hong Kong Eye Hospital, Hong Kong Special Administrative Region, China.

\textsuperscript{48} Beijing Visual Science and Translational Eye Research Institute (BERI), Beijing Tsinghua Changgung Hospital Eye Center, School of Clinical Medicine, Tsinghua Medicine, Tsinghua University, Beijing, China.

\textsuperscript{49} Institute of High Performance Computing (IHPC), Agency for Science, Technology and Research (A*STAR), Singapore 138632, Singapore.

\vspace{0.8em}

\textsuperscript{\&}These authors contributed equally

\textsuperscript{*}Corresponding authors and contributed equally: \href{mailto:chingyu.cheng@nus.edu.sg}{chingyu.cheng@nus.edu.sg}; \href{mailto:hzfu@ieee.org}{hzfu@ieee.org}; \href{mailto:haoyuchen@cuhk.edu.hk}{haoyuchen@cuhk.edu.hk}
\par}

\clearpage

\section*{Summary}

\subsection*{Background}

Foundation models hold promise for ophthalmic imaging diagnostics, but their reliance on local data labelling and task-specific fine-tuning limits scalability and clinical use. Existing coding-free AI tools are not tailored for health care and lack confidence estimation or case retrieval. We aimed to develop and evaluate a clinician-friendly, deployment-oriented foundation model for ophthalmic disease diagnosis.

\subsection*{Methods}

We developed GlobeReady, a deployment-oriented platform powered by the RetiGlobe foundation model and local feature augmentation. RetiGlobe was pretrained in two stages: 1) self-supervised learning using DINOv2 on 38 million synthetic ophthalmic images, and 2) contrastive learning using CLIP on 475,845 real image-text pairs spanning diverse ethnicities, imaging devices, and geographic regions worldwide. We evaluate GlobeReady on 488,448 ophthalmic images, including color fundus photographs (CFPs) and optical coherence tomography scans, from multi-centres in China, Singapore, Vietnam and the UK. Prospective testing included usability assessment with 31 ophthalmologists. Exploratory analyses evaluated domain generalisability, Bayesian uncertainty quantification, out-of-distribution (OOD) detection, and feature-based case retrieval.

\subsection*{Findings}

Without any fine-tuning, GlobeReady achieved high diagnostic accuracies of 93\mdot{}9\% (95\% CI: 93\mdot{}4-94\mdot{}4)--98\mdot{}5\% (95\% CI: 98\mdot{}2-98\mdot{}7) for 11 fundus diseases based on CFPs and 87\mdot{}2\% (95\% CI: 86\mdot{}9-87\mdot{}5)--92\mdot{}7\% (95\% CI: 92\mdot{}4-92\mdot{}9) for 15 retinal diseases based on OCT scans. By leveraging training-free local feature augmentation, GlobeReady effectively mitigated domain shifts across centres and populations, with a diagnostic performance increased by 63\mdot{}3\% for CFPs, achieving average accuracies of 86\mdot{}3\% (95\% CI: 83\mdot{}7-88\mdot{}8)-96\mdot{}9\% (95\% CI: 95\mdot{}5-98\mdot{}42) in Vietnam, 73\mdot{}4\% (95\% CI: 70\mdot{}6-75\mdot{}7)-91\mdot{}0\% (95\% CI: 89\mdot{}1-92\mdot{}6) in Singapore, and 90\mdot{}2\% (95\% CI: 88\mdot{}2-92\mdot{}6)-98\mdot{}9\% (95\% CI: 98\mdot{}0-99\mdot{}6) in the UK. Bayesian confidence-quantifiable diagnosis improved CFP and OCT accuracies to 99.4\% (95\% CI: 99\mdot{}2-99\mdot{}6) and 96.2\% (95\% CI: 96\mdot{}0-96\mdot{}4), respectively, while enabling reliable detection of 62 OOD conditions across CFP and OCT. GlobeReady's feature-based retrieval aligned closely with clinician judgment (Top-10 accuracy: 0\mdot{}901 for CFPs 0\mdot{}996 for OCT). In the prospective usability testing, clinicians rated GlobeReady significantly higher than Google AutoML (System Usability Scale 83\mdot{}3 vs 31\mdot{}3; \textit{P}<0\mdot{}001), with mean satisfaction of 4\mdot{}5/5.

\subsection*{Interpretation}

GlobeReady demonstrates that a clinician-friendly, deployment-oriented foundation model can deliver high diagnostic accuracy, strong generalisability across populations, robust uncertainty-based safety, and substantial usability without requiring fine-tuning or coding skills. By achieving stable performance across multiple independent international cohorts, this approach addresses significant technical barriers to AI adoption in ophthalmology. These findings provide a technical foundation for a more accessible and scalable pathway toward clinical AI application, supporting future efforts to integrate such tools into real-world clinical workflows and prospective evaluation frameworks.

\subsection*{Funding}

Supported by Agency for Science, Technology and Research and National Medical Research Council, Singapore; Shantou Science and Technology Program, Li Ka Shing Foundation, and National Key R\&D Program, China.

\section*{Research in Context}

\subsection*{Evidence before this study}

We searched PubMed and Google Scholar up to Aug 1, 2025, using the terms ``ophthalmology'', ``foundation model'', ``fundus'', ``optical coherence tomography'', ``training-free AI'', ``fine-tuning'', ``case retrieval'', and ``clinician usability''. Recent advances in artificial intelligence (AI) have led to the development of vision foundation models for ophthalmic imaging. These models were pre-trained on large-scale collections of unlabelled color fundus photographs (CFPs) and optical coherence tomography (OCT) scans, and have shown strong performance in downstream tasks such as disease detection. However, these models typically required labeled local datasets for task-specific adaptation. The dependence on data labelling and fine-tuning introduces computational and operational barriers, such as increased resource requirements, the need for technical expertise, and constraints related to data privacy and sharing. Although commercial platforms like Google AutoML or Microsoft Azure ML Studio offer no-code AI development pipelines, they are not optimized for medical imaging tasks and are often reliant on cloud infrastructure, which poses further usability and privacy challenges in clinical settings. Furthermore, most existing AI models in ophthalmology are narrow in scope, limited to a small number of common diseases, and lack the ability to detect out-of-distribution (OOD) conditions.

\subsection*{Added value of this study}

To our knowledge, this is the first study to present a training-free, deployment-oriented ophthalmic foundation model that achieves high diagnostic performance across modalities and healthcare settings, while being immediately usable by clinicians without additional coding or fine-tuning. We present GlobeReady, powered by RetiGlobe, an ophthalmic foundation model pretrained on 38 million synthetic ophthalmic images using the self-supervised framework DINOv2, followed by multimodal contrastive learning on over 475,000 real-world image-text pairs collected across diverse ethnicities, imaging devices, and geographic regions. Without any fine-tuning, GlobeReady achieved accuracies of 93\mdot{}9\% (95\% CI: 93\mdot{}4-94\mdot{}4)--98\mdot{}5\% (95\% CI: 98\mdot{}2-98\mdot{}7) for 11 fundus diseases and 87\mdot{}2\% (95\% CI: 86\mdot{}9-87\mdot{}5)--92\mdot{}7\% (95\% CI: 92\mdot{}4-92\mdot{}9) for 15 OCT-based diseases. By applying a local feature augmentation strategy, the model demonstrated significant performance improvements by 63\% based on CFPs, across sites in Vietnam, Singapore, and the UK. Incorporating Bayesian uncertainty quantification further improved diagnostic accuracy (up to 99.4\% (95\% CI: 99\mdot{}2-99\mdot{}6) for CFPs and 96\mdot{}2\% (95\% CI: 96\mdot{}0-96\mdot{}4) for OCT), while enabling robust OOD detection across 62 conditions. GlobeReady achieved higher clinician usability ratings than Google AutoML.

\subsection*{Implications of all the available evidence}

These findings provide a technical foundation for a scalable and practical pathway toward translating foundation model research into real-world clinical impact. GlobeReady demonstrates that training-free, deployment-oriented AI can deliver accurate, generalisable, and interpretable support across diverse ophthalmic settings, without the need for fine-tuning or technical expertise. Its ability to quantify diagnostic uncertainty, adapt to local populations through feature-level augmentation, and retrieve relevant cases for differential diagnosis makes it a comprehensive decision-support tool. With a local-first architecture that enhances privacy and reduces deployment costs, this approach offers a blueprint for safe and equitable AI integration across healthcare systems, including in other image-based specialties and low-resource settings.

\section*{Introduction}

Artificial intelligence (AI) has significantly advanced medical imaging analysis\cite{r1,r2}, particularly in disease diagnosis\cite{r3,r4}. In particular, the progress of AI in medical imaging is highly relevant to clinical specialities such as radiology\cite{r5}, pathology\cite{r6}, and ophthalmology\cite{r7}.

In ophthalmology, many of the major vision-threatening eye diseases, such as diabetic retinopathy (DR), age-related macular degeneration (AMD), and glaucoma, can be effectively diagnosed by imaging with diagnostic performance on par with that of a physician\cite{r8,r9}. In fact, early detection and accurate diagnosis using AI-based imaging has already been recommended as a potential strategy for the prevention of irreversible vision loss globally\cite{r10}. The two imaging modalities in ophthalmology, color fundus photography (CFP) and optical coherence tomography (OCT), are now relatively cheap, and widely available in both general and specialized settings\cite{r11,r12}.

However, although AI models have demonstrated impressive performance in diagnosing ocular diseases the adoption and translation of AI models to actual clinical care has been slow. AI models focus narrowly on specific diagnostic tasks (e.g., screening for DR), limiting their generalisability to broader and more complex clinical scenarios. Therefore, as shown in \textbf{Fig. 1a}, applying these AI models to tasks different from their original objectives typically requires fine-tuning on the target dataset to achieve effective adaptation. Recently, ophthalmic foundation model (FM) such as RETFound\cite{r13} and VisionFM\cite{r14} have emerged. Trained on extensive unannotated ophthalmic image datasets via self-supervised learning, these models showed promising performance across a range of diseases and downstream clinical tasks.

Nevertheless, a major limitation to the adoption of FMs to specific clinical contexts typically requires data labelling and fine-tuning to accommodate varying diseases or feature distributions in localized settings \textbf{(Fig. 1b).} This fine-tuning adaptation demands high-quality labelled local datasets, substantial computational resources and expertise in AI model development, posing significant barriers for the clinicians who lack specialized expertise in AI or the capacity to undertake labor-intensive data labelling\cite{r15}. Consequently, clinicians seeking to train and apply AI models to specific diagnostic tasks often face the dual burden of labelling clinical data and transferring it to technical teams for model fine-tuning, leading to delays and inefficiencies\cite{r16,r17}. Furthermore, strict institutional data-sharing policies and administrative hurdles rooted in privacy and security concerns pose significant barriers to AI adoption in the clinical environment. Finally, existing AI models often focus on a limited range of disease categories, which can lead to poor performance when encountering unseen, out-of-distribution (OOD) diseases (so-called ``zero-shot'' performance). Such performance gaps may result in misdiagnosis or overlooked conditions, adversely impacting patient outcomes and potentially leading to misdiagnosis of uncommon but potentially life-or vision-threatening conditions. In another scenario, for the uncommon diseases, clinicians often need to refer to specific cases from extensive literature or conduct a detailed image search before making a diagnosis; however, an automatic tool to facilitate this process has not yet been developed.

To address these gaps, in this study, we developed GlobeReady, a deployment-oriented ophthalmic image analysis platform that allows clinicians to seamlessly use AI for ocular disease diagnosis, by generating confidence-quantifiable ocular disease diagnosis and retrieving cases based on specific features, all without the need for specialized AI coding technical skills. Fig. 1d-h illustrates an overview of GlobeReady's architecture and application workflow. To develop GlobeReady, we collected 488,448 ophthalmic images, including 382,527 CFPs and 105,921 OCT images from open datasets, ophthalmic literature, and online resources, representing a wide range of diseases across various ethnicities and regions around the world. At the core of GlobeReady is RetiGlobe, a foundation model specifically engineered for ophthalmic image understanding. To pre-train RetiGlobe, we first generated a large-scale synthetic dataset of 38 million ophthalmic images using a generative model. We then curated 475,845 real-world image-text pairs from globally diverse sources. We initially employed the self-supervised DINOv2 framework\cite{r18} on the synthetic dataset to enable the model to learn detailed structural and pathological features. Next, we conducted contrastive pretraining with the CLIP framework\cite{r19} on the real image-text pairs, aligning visual features with semantic context to enhance complex feature understanding. These two complementary pretraining stages endow RetiGlobe with robust and generalizable ophthalmic knowledge. Building upon RetiGlobe's capabilities, we introduced the innovative GlobeReady platform, a completely code-free and training-free tool. Clinicians can effortlessly upload local data to create customized feature repositories, performing tasks such as disease diagnosis via simple feature matching. This streamlined workflow significantly simplifies the integration and practical deployment of AI technologies into everyday clinical practice. To ensure data privacy, GlobeReady is designed for on-site local implementation. Its retrieval-based mechanism allows the system to be deployed entirely within an institution's secure internal network, which precludes the need to transmit sensitive patient data to external servers. This architecture allows institutions to maintain full data sovereignty while benefiting from the model's diagnostic capabilities.

\section*{Methods}

\subsection*{Study design}

We developed GlobeReady, a clinician-friendly, training-free ophthalmic AI platform powered by the RetiGlobe foundation model. Development comprised three stages: (1) data assembly, (2) tailored pre-training, and (3) platform integration. Model performance was assessed via retrospective internal and external validation across multi-centre datasets, out-of-distribution (OOD) detection, feature-based retrieval, and a prospective usability evaluation with practising ophthalmologists. No patient-identifiable data were used.

\subsection*{Data sources}

We assembled 382,527 colour fundus photographs (CFPs) and 105,921 optical coherence tomography (OCT) scans from 39 de-identified public datasets, ophthalmic literature, and online repositories spanning diverse ethnicities, imaging devices, and regions \textbf{(Table S1).} To enhance representation robustness and domain diversity, we generated 38 million synthetic ophthalmic images (25 million CFPs; 13 million OCT) using FundusGAN\cite{r20}. The base real images feeding FundusGAN were sourced from 39 globally diverse public datasets comprising 382,527 CFPs and 105,921 OCT scans, which encompass 414 distinct clinical categories as detailed in \textbf{the Appendix (Disease Information in FundusGAN Training Datasets). }The synthetic images were generated randomly to maximize visual diversity, and a simple automated filter was applied after generation to remove outputs that were completely blank or clearly malformed. Importantly, these synthetic images were utilized exclusively for first-stage self-supervised pretraining to enrich the feature space. Separately, we curated 475,845 ophthalmic image-text pairs (374,424 CFP--text\cite{r21}; 101,421 OCT--text) from the collected datasets. While inherent annotation noise may exist in these large-scale pretraining pairs, they were primarily utilized to capture broad semantic associations, whereas diagnostic performance was rigorously validated on separate, expert-annotated evaluation sets to ensure accuracy.

\subsection*{Model development}

The GlobeReady framework is built upon a large Vision Transformer (ViT-L/16)\cite{r22} backbone (approximately 307 million parameters). We injected Low-Rank Adaptation (LoRA)\cite{r23} modules (rank r=8) into the query and value projections for efficient adaptation while the DINOv2-pretrained weights remained frozen, adding only 0.8 million trainable parameters (0.26\%). Visual and textual representations, the latter from a Bio-ClinicalBERT text encoder, were aligned in a shared 512-dimensional embedding space, optimised via a symmetric cross-entropy loss with an evidential uncertainty objective. Full architecture, hyperparameters, augmentation strategies, and training schedules are detailed in \textbf{Appendix Section 1.}

GlobeReady was pretrained in two stages. First, self-supervised DINOv2\cite{r18} was applied to the 38 M synthetic images to learn structural and topological ophthalmic features. Second, contrastive multimodal pre-training (CLIP\cite{r19}) on 475,845 image-text pairs aligned visual and textual representations. The image-text pairs were assembled from public datasets with diagnostic labels, ophthalmic literature, and curated online repositories, and underwent a two-stage expert validation by ophthalmologists and senior retinal specialists; full provenance and validation procedures are detailed in \textbf{Appendix Section 1}.

\subsection*{Platform integration and inference}

GlobeReady is a code-free platform supporting: (1) disease diagnosis, (2) confidence-quantifiable predictions, and (3) feature-based case retrieval. Query images are encoded into embeddings and matched to a reference feature library; predictions inherit the label(s) of the most similar reference exemplars. This retrieval-based design removes the need for fine-tuning and enables rapid deployment across sites. Specifically, the user interaction follows a streamlined three-step workflow:

\begin{itemize}
  \item Data Input: Users upload ophthalmic images (e.g., CFP or OCT) to the platform.
  \item Automated Analysis: GlobeReady functions as a general diagnostic assistant, rather than requiring users to manually specify a task. The model attempts to match input images to known disease patterns within its reference space.
  \item Comprehensive Output: The user receives diagnostic predictions along with confidence scores and retrieval-based reference cases.
\end{itemize}
Further details and UI are shown in the \textbf{Appendix Guidelines and Appendix Videos}.

\subsection*{Retrospective internal validation}

For CFPs, the JLHW11 dataset (19 478 images; 10 retinal diseases + normal) was split into a reference subset (11 670) and a test subset (7 808) \textbf{(Table S2).} For OCTs, the JSIEC-OCT15 dataset (102 893 scans; 15 diseases) was split into a reference set (59 590) and a test set (43 303) \textbf{(Table S3).} Imaging devices and acquisition sites are summarized in \textbf{Table S4.}

\subsection*{Retrospective external validation}

For CFPs, we evaluated adaptive ocular disease diagnosis at seven independent centres across China, Singapore, the UK, and Vietnam using site-specific local feature augmentation:

\begin{itemize}
  \item \textbf{China: }Guangxi Jingliang Eye Hospital (GJEH; NIDEK AFC-100), Qingdao Central Hospital (QCH; Canon CX-1), and Puning People's Hospital (PNH; Topcon TRC-NW8) \textbf{(Tables S5-S7).}
  \item \textbf{Singapore: }SEED cohort\cite{r24} (Malay, Indian, Chinese ethnic groups)\textbf{ (Table S8).}
  \item \textbf{UK:} UKDR screening programme \textbf{(Table S9).}
  \item \textbf{Vietnam:} Binh Dinh Eye Hospital (BDEH) dataset \textbf{(Table S10).}
\end{itemize}
For OCTs, we used WAEH-OCT13 (Wuhan Aier Eye Hospital), Zeiss-OCT11 (JSIEC; Zeiss Cirrus HD-OCT 5000), OCTDL\cite{r67} and OCTID\cite{r68} as detailed in \textbf{Tables S11-S14)}. To harmonize diverse multi-centre data, all site-specific labels were systematically mapped to a predefined unified disease taxonomy and clinical grading framework prior to evaluation, as in \textbf{Appendix Section 2.}

\subsection*{OOD detection datasets}

For CFP OOD detection, CFPOOD49 comprises 7,257 images from 49 disease categories absent from JLHW11, plus 1,066 low-quality CFPs (Fig. 4; Table S4). For OCTs, OCT-OOD13 includes 2,594 scans across 13 unseen categories, plus 2,016 low-quality OCTs (Fig. 4; Table S4).

\subsection*{Feature-based retrieval evaluation}

CFP retrieval used 168 query CFPs (12 diseases; BDEH; Canon CR2 AF and Crystalvue NFC-600) against JLHW11 + CFPOOD49 + CFP700K (747 237 unlabeled CFPs). OCT retrieval used 150 query OCTs (JSIEC; Topcon DRI Triton) against 34,384 unlabeled OCTs from Wuhan Aier Eye Hospital (same device). Retrieval protocols and clinician adjudication are in Appendix Section 4.

\subsection*{Confidence quantification}

We implemented Monte Carlo\cite{r25} dropout for Bayesian uncertainty\cite{r26} during inference. For each reference exemplar, 100 stochastic passes yielded an embedding distribution; query similarity across passes produced a confidence score. To ensure a real-world assessment of robustness, a fixed data-driven threshold $\Theta$ was pre-specified by maximizing Youden's index\cite{r27} on a hold-out development subset distinct from the validation cohorts (equations and optimization procedure in \textbf{Appendix Section 5}). This threshold was then independently applied to all validation cohorts without further optimization, ensuring that the reported performance reflects the model's generalisability to unseen data distributions.

\subsection*{Performance Metrics}

Diagnostic performance was evaluated using a comprehensive suite of clinical metrics, including Top-k accuracy, sensitivity, specificity, positive predictive value (PPV), and negative predictive value (NPV). Specifically, Top-k accuracy was defined as the proportion of cases in which the reference diagnosis was included within the k most confident predictions. To account for statistical uncertainty, 95\% confidence intervals (CIs) for all key metrics were generated across all evaluation cohorts using bootstrapping methods with 1,000 iterations. The normality of paired differences was confirmed (Shapiro-Wilk p>0\mdot{}05) before applying two-sided paired t-tests were employed to assess the significance of performance differences between platforms, with \textit{P }< 0\mdot{}05 considered statistically significant. This study was reported in accordance with the \textbf{Appendix STARD AI checklist}.

\subsection*{Prospective questionnaire-based usability evaluation}

We conducted a prospective questionnaire-based usability evaluation to assess the clinician-facing usability and perceived workflow utility of GlobeReady. A total of 31 ophthalmologists from China, Singapore, Vietnam, Australia, and the United States participated in the evaluation, including 5 interns, 7 residents, 15 attending ophthalmologists/consultants, and 4 chief/director-level clinicians, with clinical experience ranging from 1 to 30 years. Participants were provided with access to the GlobeReady platform, written guidelines, and demonstration videos covering its three core functions: ophthalmic disease diagnosis, confidence-quantifiable diagnosis, and feature-based case retrieval. They were instructed to test the platform using the provided evaluation images and/or their own ophthalmic images following the standardized usage guidelines. Usability was assessed using a structured questionnaire incorporating the System Usability Scale (SUS)\cite{r29} and additional Likert-scale\cite{r30} items evaluating perceived effectiveness, helpfulness, ease of use, satisfaction, time-saving potential, and willingness to use the platform in future work. To benchmark GlobeReady against an existing code-free machine learning platform, 13 ophthalmologists also evaluated Google AutoML\cite{r28} using the same questionnaire, enabling within-participant comparison between the two platforms. Questionnaire results were summarized as mean $\pm$ standard deviation. Comparisons between GlobeReady and AutoML were performed using a linear mixed-effects model, with platform type as the fixed effect and participant identity as a random effect to account for repeated measurements among ophthalmologists who evaluated both platforms. Two-sided \textit{P} values <0.05 were considered statistically significant. Details of the questionnaire design, scoring, guidelines, and demonstration materials are provided in \textbf{Appendix Section 6.}

\subsection*{Ethics}

This study was approved by the Joint Shantou International Eye Center (JSIEC) Institutional Review Board and adhered to the principles of the Declaration of Helsinki. All datasets used in this study were made de-identified before computational analysis and model development.

\section*{Results}

\subsection*{Ocular disease diagnosis by GlobeReady}

With GlobeReady's built-in ocular disease diagnosis feature (Fig. 1b), GlobeReady demonstrated exceptional performance across multiple datasets involving diverse disease categories and imaging modalities. Specifically, it achieved a Top-1 accuracy of 93\mdot{}9\% (95\% CI: 93\mdot{}4-94\mdot{}4), sensitivity of 91\mdot{}6\% (95\% CI: 90\mdot{}8-92\mdot{}5), specificity of 99\mdot{}4\% (95\% CI: 99\mdot{}3-99\mdot{}4), PPV of 93\mdot{}6\% (95\% CI: 92\mdot{}9-94\mdot{}3), and NPV of 99\mdot{}4\% (95\% CI: 99\mdot{}3-99\mdot{}4) on a dataset comprising 7,808 CFPs covering 11 disease categories from 4 hospitals in China (termed as JLHW11 dataset; \textbf{Fig. 2a, Appendix Fig. S1a}), which includes CFPs of 10 retinal disorders and normal conditions (\textbf{Table S2}). Additionally, our model achieved Top-3 and Top-5 accuracies of 97\mdot{}6\% (95\% CI: 97\mdot{}2-97\mdot{}8) and 98\mdot{}5\% (95\% CI: 98\mdot{}2-98\mdot{}7), respectively, with corresponding sensitivities of 96\mdot{}0\% (95\% CI: 95\mdot{}4-96\mdot{}7) and 97\mdot{}3\% (95\% CI: 96\mdot{}8-97\mdot{}8), specificities of 97\mdot{}9\% (95\% CI: 97\mdot{}8-98\mdot{}0) and 96\mdot{}6\% (95\% CI: 96\mdot{}5-96\mdot{}8), PPVs of 81\mdot{}2\% (95\% CI: 80\mdot{}3-82\mdot{}3) and 73\mdot{}4\% (95\% CI: 72\mdot{}5-74\mdot{}5), and NPVs of 99\mdot{}7\% (95\% CI: 99\mdot{}7-99\mdot{}8) and 99\mdot{}8\% (95\% CI: 99\mdot{}8-99\mdot{}8). Notably, the Top-5 sensitivities ranged from 88\mdot{}2\% (95\% CI: 84\mdot{}1-91\mdot{}8) to 99\mdot{}9\% (95\% CI: 99\mdot{}7-100\mdot{}0) across all conditions (\textbf{Table S18}), including macular, optic nerve head--related, retinal vascular, and other retinal diseases.

We further evaluated GlobeReady's diagnostic performance using an OCT dataset curated by the Joint Shantou International Eye Center spanning 15 types of ocular conditions (JSIEC-OCT15 dataset;\textbf{ Fig. 2a, Appendix Fig. S1b; Table S21-S24}). The model demonstrated promising results, achieving Top-1, Top-3, and Top-5 accuracies of 87\mdot{}2\% (95\% CI: 86\mdot{}9-87\mdot{}5), 90\mdot{}9\% (95\% CI: 90\mdot{}6-91\mdot{}1), and 92\mdot{}7\% (95\% CI: 92\mdot{}4-92\mdot{}9), respectively. The corresponding average sensitivities were 86\mdot{}3\% (95\% CI: 85\mdot{}9-86\mdot{}6), 89\mdot{}8\% (95\% CI: 89\mdot{}5-90\mdot{}1), and 91\mdot{}7\% (95\% CI: 91\mdot{}4-91\mdot{}9), with specificities of 99\mdot{}1\% (95\% CI: 99\mdot{}1-99\mdot{}1), 98\mdot{}7\% (95\% CI: 98\mdot{}7-98\mdot{}7), and 98\mdot{}5\% (95\% CI: 98\mdot{}4-98\mdot{}5), positive predictive values of 86\mdot{}0\% (95\% CI: 85\mdot{}6-86\mdot{}4), 82\mdot{}0\% (95\% CI: 81\mdot{}6-82\mdot{}5), and 79\mdot{}8\% (95\% CI: 79\mdot{}3-80\mdot{}2), and negative predictive values of 99\mdot{}1\% (95\% CI: 99\mdot{}1-99\mdot{}1), 99\mdot{}3\% (95\% CI: 99\mdot{}3-99\mdot{}4), and 99\mdot{}5\% (95\% CI: 99\mdot{}4-99\mdot{}5). Notably, the Top-5 specificities ranging from 71\mdot{}8\% (95\% CI: 69\mdot{}8-73\mdot{}7) to 99\mdot{}5\% (95\% CI: 99\mdot{}3-99\mdot{}7) across all 15 conditions.

An ablation study confirmed the contribution of large-scale synthetic pretraining. When the 38 million synthetic images were omitted from the first pretraining stage, Top-1 accuracy dropped from 93.9\% (95\% CI: 93\mdot{}4-94\mdot{}4) to 74.2\% (95\% CI: 73\mdot{}2-75\mdot{}2) on the JLHW11 dataset and from 87.2\% (95\% CI: 86\mdot{}9-87\mdot{}5) to 56.7\% (95\% CI: 56\mdot{}2-57\mdot{}7) on the JSIEC-OCT15 dataset \textbf{(Appendix Fig. S13).}

\subsection*{Adaptive ocular disease diagnosis by local retrieval augmented GlobeReady}

AI models often necessitate fine-tuning to adapt to varying tasks or domains, such as different clinics or geographical regions, due to variations in equipment, ethnicity, and disease prevalence. Drawing inspiration from retrieval augmented generation (RAG) techniques in large language models (LLMs)\cite{r31,r32}, GlobeReady addresses this challenge by enhancing the base reference features with local reference features, eliminating the need for fine-tuning \textbf{(Fig. 1c).} We used features extracted from the reference subset of the JLHW11 dataset from China as the CFP global reference feature library and used diverse external datasets served as the sources for local reference features. Local reference features are embedding representations extracted via the frozen RetiGlobe encoder from site-specific reference data, encoding both anatomical structures and population-specific pathological patterns unique to the local deployment site. Evaluation was conducted on independent local test sets that were strictly excluded from the construction of the local reference libraries. For each test sample GlobeReady identifies the nearest neighbor within an integrated global-local embedding space using cosine similarity (Eqs. (1)-(2), \textbf{Appendix Section 3}); the query inherits the label of this nearest reference, and candidate diagnoses are ranked by similarity for Top-n reporting. This non-parametric matching mechanism enables the model to leverage site-specific pathological patterns and imaging artefacts remaining frozen thereby significantly enhancing diagnostic performance. As illustrated in \textbf{Fig. 2b,} in all the participating centers, diagnostic performance across all participating centres improved when local features were augmented, compared to when they were not, particularly in the centres outside China (\textbf{Appendix Fig. S2}, \textbf{Table S25-S56}).

We further evaluated GlobeReady's performance across three clinical centres in different countries. With local retrieval augmentation, GlobeReady's diagnostic performance improved by 63\mdot{}3\% on average for CFP and 17\mdot{}8\% on average for OCT. Specifically, at the Singapore National Eye Centre, its performance gains across multi-ethnic populations in the Singapore Epidemiology of Eye Diseases (SEED) Study\cite{r24}, with Top-1 to Top-5 accuracies improved by 28\mdot{}3-31\mdot{}1\%, reaching to 73\mdot{}4\% (95\% CI: 70\mdot{}6-75\mdot{}7)-91\mdot{}0\% (95\% CI: 89\mdot{}1-92\mdot{}6) (\textit{P}$\leq$ 0.001 compared to before local retrieval augmentation; \textbf{Fig.2b, Appendix Fig. S2d, Table S37-S40}). At Binh Dinh Eye Hospital in Vietnam, accuracies rose from 51\mdot{}0--72\mdot{}7\% to 86\mdot{}3\% (95\% CI: 83\mdot{}7-88\mdot{}8)-96\mdot{}9\% (95\% CI: 95\mdot{}5-98\mdot{}42) (\textit{P} $\leq$ 0.002), with similar gains observed in sensitivities\textbf{ (Fig. 2b, Appendix Fig. S2e, Table S41-S44)}. Likewise, at the Department of Eye and Vision Sciences, University of Liverpool in the United Kingdom, GlobeReady's diagnostic performance improved significantly by 76\mdot{}9-168\mdot{}5\% following local feature augmentation, with accuracies rising from 33\mdot{}6\% (95\% CI: 30\mdot{}3-37\mdot{}3)-55\mdot{}9\% (95\% CI: 51\mdot{}8-59\mdot{}6) to 90\mdot{}2\% (95\% CI: 88\mdot{}2-92\mdot{}6)-98\mdot{}9\% (95\% CI: 98\mdot{}0-99\mdot{}6) \textbf{( \textit{P}}\textbf{ $\leq$ 0.002, Fig. 2b, Appendix Fig. S2f, and Table S45-S48)} for CFPs collected from a diabetic retinopathy (DR) screening program (referred as UKDR dataset thereafter).

We further evaluated GlobeReady's performance in identifying ocular diseases in OCT images with local retrieval augmented diagnosis, using features from the reference subset of JSIEC-OCT15 as the base OCT reference feature library. In the WAEH-OCT13 dataset, which encompasses 12 ocular diseases along with normal conditions, Top-1, Top-3, and Top-5 accuracies increased from ranges of 81\mdot{}2\% (95\% CI: 79\mdot{}7-82\mdot{}4)--89\mdot{}0\% (95\% CI: 87\mdot{}8-90\mdot{}0) to 89\mdot{}0\% (95\% CI: 88\mdot{}0-90\mdot{}2)--92\mdot{}5\% (95\% CI: 91\mdot{}5-93\mdot{}4) (\textit{P} $\leq$ 0\mdot{}001), with sensitivities improving to 86\mdot{}6\% (95\% CI: 84\mdot{}9-88\mdot{}2)--91\mdot{}3\% (95\% CI: 89\mdot{}8-92\mdot{}3) \textbf{(Fig. 2b, Appendix Fig. S2g, Table S49-S52).} Similarly, in the Zeiss-OCT11 dataset, accuracies rose from 61\mdot{}3\% (95\% CI: 59\mdot{}0-63\mdot{}6)--77\mdot{}2\% (95\% CI: 75\mdot{}2-79\mdot{}3) to 83\mdot{}7\% (95\% CI: 81\mdot{}7-85\mdot{}6)--94\mdot{}9\% (95\% CI: 93\mdot{}7-96\mdot{}0) (\textit{P} $\leq$ 0\mdot{}001) following the augmentation of local features, accompanied by comparable improvements in sensitivities \textbf{(Fig. 2b, Appendix Fig. S2h, Table S53-S56).}

To isolate the contribution of this augmentation, we compared the combined ``Global + Local'' retrieval against a ``Local-only'' baseline across all cohorts. The results consistently demonstrated that the integrated retrieval space outperformed local-only retrieval\textbf{ (Appendix Fig. S3, Table S25-S56),} confirming that the performance gains arise from the synergistic interaction between global knowledge and local adaptation rather than the mere inclusion of local data.

Furthermore, we compared our training-free retrieval approach with specialized models including RETFound and VisionFM under a few-shot full fine-tuning protocol. As illustrated in \textbf{Appendix Fig. S4}, despite these models undergoing full parameter updates to adapt to local distributions, GlobeReady's non-parametric configuration consistently maintained higher overall accuracy and sensitivity across multiple benchmarks. This indicates that our specialized ocular representation combined with the local augmentation strategy captures clinical features more effectively than traditional weight-update methods in data-constrained scenarios. These findings underscore that GlobeReady not only bypasses the operational burdens of model retraining but also provides a more robust alternative for clinical deployment.

\subsection*{Confidence-quantifiable ocular disease diagnosis with Bayesian GlobeReady}

Providing diagnostic results with a reliability assessment is crucial for deploying AI models in clinical practice. Therefore, we integrated Bayesian uncertainty evaluation into our GlobeReady platform\textbf{ (Fig. 1d).} This feature improved GlobeReady's performance on the JLHW11 dataset, increasing Top-1 accuracy from 93\mdot{}9\% (95\% CI: 93\mdot{}3-94\mdot{}4) to 94\mdot{}9\% (95\% CI: 94\mdot{}4-95\mdot{}3) (\textit{P} = 0\mdot{}001) and Top-1 sensitivities from 91\mdot{}6\% (95\% CI: 89\mdot{}2-94\mdot{}1) to 92\mdot{}1\% (95\% CI: 89\mdot{}8-94\mdot{}5) with \textit{P} = 0\mdot{}007 (\textbf{Fig. 2c, Table S57}). Similar improvements were noted in the JSIEC-OCT15 dataset, with Top-1 accuracy rising from 87\mdot{}2\% (95\% CI: 86\mdot{}9-87\mdot{}5) to 88\mdot{}2\% (95\% CI: 87\mdot{}9-88\mdot{}5) (\textit{P} < 0\mdot{}001) and sensitivity reaching 87\mdot{}0\% (95\% CI: 85\mdot{}8-88\mdot{}3) with \textit{P} < 0\mdot{}001 (\textbf{Fig. 2c, Table S58}). To further quantify the reliability of this assessment, we performed an Uncertainty-Error ROC analysis, utilizing the uncertainty value as a continuous variable to evaluate the system's ability to identify potential diagnostic errors (\textbf{Appendix Fig. S6}). The uncertainty module achieved an AUC of 0\mdot{}922 on the JLHW11 dataset and 0\mdot{}806 on the JSIEC-OCT15 dataset, demonstrating its robust capacity for error detection.

Furthermore, implementing a confidence threshold to flag cases for further clinical review would reduce misdiagnosis risk. Post-flagging, GlobeReady's Top-1 accuracy and sensitivities on the JLHW11 dataset improved to 99\mdot{}4\% (95\% CI: 99\mdot{}2-99\mdot{}6) and 98\mdot{}5\% (95\% CI: 97\mdot{}1-99\mdot{}8) (\textit{P} = 0\mdot{}001), respectively, with sensitivities across categories ranging from 91\mdot{}2\% (95\% CI: 83\mdot{}9-98\mdot{}6) to 100\% (95\% CI: 100\mdot{}0-100\mdot{}0). For OCT images, similar strategies enhanced Top-1 accuracy to 96\mdot{}2\% (95\% CI: 96\mdot{}0-96\mdot{}4) and sensitivity to 94\mdot{}5\% (95\% CI: 93\mdot{}4-95\mdot{}5) on the JSIEC-OCT15 dataset (\textit{P} < 0\mdot{}001).

We fine-tuned two other foundation models, RETFound\cite{r13} and VisionFM\cite{r14}, and then compared their diagnostic performance with our GlobeReady. GlobeReady showed enhanced performance in Top-3, Top-5, and confidence-thresholded accuracy and sensitivities without fine-tuning (\textit{P} $\leq$ 0\mdot{}001), and achieved comparable or higher Top-1 across both JLHW11 and JSIEC-OCT15 datasets (\textbf{Appendix Fig. S5}). \textbf{As shown in Fig. 3,} we conducted t-SNE visualizations of feature embeddings extracted by the pretrained weights of RETFound\cite{r13}, VisionFM\cite{r14}, and GlobeReady on the JLHW11 and JSIEC-OCT15 datasets, without any fine-tuning. Each point is color-coded by disease category. Compared to RETFound\cite{r13} and VisionFM\cite{r14}, GlobeReady exhibited more distinct inter-class separation and well-defined cluster boundaries, which was further validated by quantitative clustering metrics. On the JLHW11 dataset, GlobeReady achieved a Silhouette Score (SC) of 0.048, a Davies-Bouldin Index (DBI) of 3.9532 and a Calinski-Harabasz Index (CHI) of 263.36, outperforming both RETFound\cite{r13} (SC=-0.159; DBI: 16.562; CHI: 86.530) and VisionFM\cite{r14} (SC=0.055; DBI: 3.842; CHI: 225.000). Similarly, On the JSIEC-OCT15 dataset, GlobeReady reached the highest SC (0.061) and CHI (1788.670), alongside the lowest DBI (3.174), indicating more robust disease-specific encoding compared to VisionFM (SC: 0.047; DBI: 3.387; CHI: 1555.22) and RETFound (SC: -0.117; DBI: 9.071; CHI: 212.590), reflecting stronger disease-specific feature encoding. We note that while certain disease classes remain relatively dispersed this pattern likely reflects intrinsic pathological heterogeneity where single diagnostic categories encompass diverse morphological subtypes and severity stages. Furthermore, overlapping visual features across related conditions and device-driven variability in acquisition protocols contribute to the continuous nature of the embedding space. Rather than forcing artificial clusters GlobeReady preserves these nuanced transitions to capture the complex spectrum of pathological changes which is essential for a retrieval-based framework. These results demonstrate the competitive representational capacity of GlobeReady and its ability to represent a broad spectrum of retinal diseases by reflecting real-world clinical complexity.

\subsection*{Retrieval-based benchmarking evaluations with GlobeReady}

To elucidate GlobeReady's advantages over current FMs, we conducted a rigorous comparative analysis within a unified framework. Beyond initial comparisons with RETFound and VisionFM, we expanded the benchmarking scope to include high-capacity general-purpose models, specifically DINOv2\cite{r18} and DINOv3\cite{r69}. To ensure a fair assessment of underlying representational capacities, all baseline models were evaluated under two distinct protocols: training-free retrieval pipeline and a few-shot linear probing strategy. Furthermore, we extended our evaluation to include two widely recognized public datasets, OCTDL\cite{r67} and OCTID\cite{r68}, to satisfy the requirement for transparent and reproducible benchmarking.

As illustrated in \textbf{Appendix Fig. S7}, quantitative results demonstrate that GlobeReady consistently maintains highly competitive performance across both internal and public datasets. Notably, our analysis reveals that retrieval-based methods (indicated by light colors) consistently outperform few-shot linear probing (indicated by dark colors) across all tested backbones. This suggests that when local labeled data is extremely sparse, parameter-based adaptation like linear probing is prone to sub-optimal convergence or overfitting, whereas GlobeReady's non-parametric retrieval leverages the foundational feature space more effectively. While DINOv2 and DINOv3 possess robust general features, they yielded relatively lower diagnostic accuracy in most ophthalmic tasks, highlighting the necessity of domain-specific pretraining for clinical applications. Interestingly, although RETFound demonstrated greater representational depth than general-purpose models on standardized public OCT datasets, its performance deteriorated significantly on the JLHW11 cohort, falling even below the general-purpose baselines. This discrepancy elucidates a pivotal limitation of contemporary specialized foundation models: their efficacy is heavily contingent upon pre-training distributions, necessitating task-specific fine-tuning when confronted with significant variations in hardware or population cohorts. Conversely, GlobeReady consistently achieved robust outcomes across all benchmarks, demonstrating an innate adaptability that effectively bridges the chasm between specialized medical expertise and cross-centre robustness.

\subsection*{Open set anomaly detection with Bayesian GlobeReady}

To assess GlobeReady's reliability in real-world clinical settings where patients may present with previously unseen conditions, we leveraged the Bayesian uncertainty framework to perform open set anomaly detection. We constructed a new evaluation dataset (termed CFPOOD49), containing 7,257 fundus images across 49 disease categories entirely absent from the original JLHW11 reference set \textbf{(Fig. 4a).} Using features from the JLHW11 reference subset as the base CFP feature library, GlobeReady identified anomaly samples by computing predictive uncertainty scores for each input. Samples were flagged as anomalies when their uncertainty exceeded the predefined threshold, signaling that the input fell outside the model's known diagnostic space. This confidence threshold strategy is intended as a critical clinical safety-oriented function ensuring that rare or novel conditions are triaged for expert clinician review rather than automated decision-making.

GlobeReady accurately flagged the anomaly samples in the CFPOOD49 dataset, achieving detection rates ranging from 66\mdot{}7\% to 100\mdot{}0\%. Several conditions, including choroidal osteoma and multiple evanescent white-dot syndromes, reached perfect detection (100\%). Most diseases showed strong detection rates (85-95\%), while a few, such as Purtscher retinopathy, were lower due to small sample sizes. Notably, epiretinal membrane, the largest class with 2,097 images, achieved 90\mdot{}2\%. Additionally, GlobeReady demonstrated robust performance on a separate dataset of 1,066 low-quality fundus images, reaching a 94\mdot{}6\% anomaly detection rate using a confidence threshold strategy.

We further validated GlobeReady's ability to detect unfamiliar disease categories in OCT images. A new evaluation dataset, OCT-OOD13, was curated, comprising 2,594 OCT scans across 13 disease categories not present in the JSIEC-OCT15 reference set \textbf{(Fig. 4b)}. Using features from JSIEC-OCT15 as the base OCT feature library, GlobeReady flagged most samples with unseen conditions from OCT-OOD13 dataset, achieving detection rates ranging from 66\mdot{}7\% to 100\mdot{}0\%. Perfect detection was observed for Stargardt disease and optic disc pit maculopathy (100\mdot{}0\%), while conditions such as silicone oil-filled eyes, punctate inner choroidopathy, and choroidal hemangioma also achieved high detection rates (>94\%). Moderate-to-high performance was observed for chronic retinal vein occlusion, focal choroidal excavation, and vitreoretinal lymphoma (83-88\%), demonstrating GlobeReady's strong generalisability across diverse OCT abnormalities. To further assess robustness against image quality degradation, we tested the model on 2,016 low-quality OCT scans, achieving a 92\mdot{}8\% detection rate using confidence-threshold filtering.

\subsection*{Feature-based case retrieval with GlobeReady}

GlobeReady facilitates rapid retrieval of similar cases without the need for fine-tuning \textbf{(Fig. 1e).} We assessed its performance using an additional set of 168 CFPs, across 12 ocular diseases, collected from Vietnam \textbf{(Appendix Fig. S8a)} to retrieve similar cases from the JLHW11 and CFPOOD49 datasets. Retrieval performance varied across diseases, with Top-1 to Top-10 sensitivities differing by condition \textbf{(Appendix Fig. S8b).} The highest Top-1 sensitivity was observed for epiretinal membrane (0\mdot{}968), followed by sunset-glow fundus (0\mdot{}889). Central serous chorioretinopathy and hypertensive retinopathy achieved 0\mdot{}795 and 0\mdot{}750, respectively. Rare conditions such as macular coloboma and optic neuritis initially showed poor Top-1 sensitivity, but improved to 1\mdot{}0 when considering Top-3 or Top-10 results.

To evaluate GlobeReady clinically, we conducted a retrieval study using these 168 CFPs as queries against a large-scale reference database (CFP700K) containing 747,237 unlabeled CFPs sourced from multiple hospitals and diverse imaging devices. GlobeReady returned the Top-10 most similar images per query, yielding 1,680 retrieved images. These were reviewed by three experienced ophthalmologists via an online platform we developed\textbf{ (Appendix Fig. S9).} The assessments showed average accuracy rates of 0\mdot{}522 for Top-1, 0\mdot{}714 for Top-3, 0\mdot{}792 for Top-5, and 0\mdot{}901 for Top-10, demonstrating GlobeReady's high reliability in retrieving clinically relevant cases \textbf{(Fig. 5a),} especially when multiple top results are considered. Additionally, GlobeReady's OCT retrieval performance was tested using 150 OCT images from JSIEC as queries against 34,384 unlabeled OCT scans from Wuhan Aier Eye Hospital. The retrieval results, assessed by three ophthalmologists via our platform\textbf{ (Appendix Fig. S10)}, showed efficient and accurate retrieval of highly similar OCT cases without model fine-tuning \textbf{(Fig. 5b).} The average Top-1 accuracy was 0.920, closely aligning with clinical judgment, and improved to 0\mdot{}984 for Top-3, 0\mdot{}993 for Top-5, and 0\mdot{}996 for Top-10.

\subsection*{Questionnaire evaluations for GlobeReady platform}

To conduct an exploratory evaluation of the usability and user-friendliness of the GlobeReady platform, we compared its performance with another coding-free platform, Google's AutoML\cite{r28}. A total of 31 ophthalmologists from China, Singapore, Vietnam, Australia, and the United States were invited to complete a questionnaire based on their experience. 13 of the participating ophthalmologists used both platforms and the others only used the GlobeReady platform. The difference of the performance between two platforms were analyzed through linear mixed model, accounting for the within-subject correlation among participants who evaluated both platforms. The characteristics of the participating ophthalmologists were shown in \textbf{Table 1}. The questionnaire consisted of the system usability scale (SUS)\cite{r30}, effectiveness, helpfulness, ease of use, and satisfaction (\textbf{Appendix Questionnaire}). Among the 31 participating ophthalmologists, thirteen (41.9\%) used both fundus and OCT images, nine (29.0\%) used only fundus images, and nine (29.0\%) used only OCT images. As shown in \textbf{Table 2}, GlobeReady achieved a significantly higher average SUS (System Usability Scale) score of 83.3$\pm$11.2 compared to AutoML's 31.3$\pm$12.1 (P<0\mdot{}001). This result indicated that GlobeReady reached a level of ``excellent'' usability, while AutoML\cite{r28} reached a level of ``awful'' usability\cite{r30}. Across other evaluated dimensions using a 5-point Likert scale, GlobeReady consistently outperformed AutoML. The average Likert scores for GlobeReady were 4.1 $\pm$ 0\mdot{}7 for effectiveness, 4.4 $\pm$ 0\mdot{}9 for helpfulness, 4.4 $\pm$ 0\mdot{}6 for ease of use, and 4\mdot{}5 $\pm$ 0.7 for overall satisfaction, indicating a high level of user acceptance (\textbf{Table 2}). The clinicians specifically noted that GlobeReady could reduce the time required for clinical diagnosis (Likert score = 4.1$\pm$0.9), while AutoML offered limited time-saving benefits (Likert score = 2\mdot{}1 $\pm$ 0\mdot{}9) and often required the users to invest additional effort to interpret or correct results (converted Likert score = 2\mdot{}4 $\pm$ 0\mdot{}9). Furthermore, the clarity of GlobeReady's user interface was rated highly (Likert score = 4.4 $\pm$ 0\mdot{}6), in contrast to AutoML's interface, which was considered less intuitive (Likert score = 2\mdot{}5 $\pm$ 0\mdot{}8). Importantly, most of the ophthalmologists expressed a clear preference for using GlobeReady in future clinical tasks (Likert score = 4.5 $\pm$ 0\mdot{}7), compared to a much lower preference for AutoML (Likert score = 2\mdot{}2 $\pm$ 0\mdot{}7). These results demonstrate that GlobeReady offers a highly user-friendly experience tailored to the needs of clinicians. In addition to usability, several participants also validated GlobeReady's diagnostic performance in their local settings, as shown previously (SEED, BDEH, and Zeiss-OCT11 in \textbf{Fig. 2b}).

\subsection*{Model interpretation}

To investigate the attribution patterns of GlobeReady's pretrained weights, without task-specific fine-tuning, across ocular imaging modalities, we employed AGFVisualization\cite{r33}, a saliency mapping technique. As shown in \textbf{Fig. 6}, the model intrinsically localized disease-specific features in CFPs, such as the retinal vascular abnormalities for branch retinal vein occlusion and the macular area for wet AMD. Similarly, OCT analysis demonstrated precise attention to pathological structures, such as the maculoschisis with posterior staphyloma and chorioretinal thinning in myopic traction maculopathy (\textbf{Fig. 7}). These results indicate the model's ability to capture clinically relevant pathological regions. To further assess interpretability across a wider range of conditions, we generated heatmaps for additional diseases \textbf{(Fig. S11-S12).} In most cases, such as diabetic retinopathy, central retinal vein occlusion, and retinal detachment, the model focused on expected pathological structures. However, we also observed instances where attention patterns were less localized or more diffuse due to ambiguous features (\textbf{Appendix Fig. S11g-h, Fig. S12g-h}). For instance, in \textbf{Fig. S11g-h}, the model misdirected its attention to irrelevant regions for glaucoma and Vogt-Koyanagi-Harada disease, leading to erroneous predictions. Similarly, \textbf{Fig. S12g-h} shows less localized attention for neovascular AMD and more diffuse attention for dry AMD. These visualization results provide mechanistic evidence for GlobeReady's robust feature-matching performance in typical cases, while also highlighting potential limitations in atypical or complex presentations.

\section*{Discussion}

A critical limitation to the adoption of FMs in medicine is that for models to be useful in specific clinical contexts, data labeling and fine-tuning are typically required to accommodate local feature distributions. This process requires expertise in AI model development and coding, posing significant barriers for the average clinician who may lack specialized technical skills or the capacity for labor-intensive data annotation. We present GlobeReady, a deployment-oriented AI platform that enables direct deployment without requiring coding expertise or model fine-tuning \textbf{(Appendix guidelines and Appendix videos).} By integrating DINOv2\cite{r18} with a CLIP-based pretraining framework\cite{r19}, GlobeReady acquires robust and generalizable feature representations that enable immediate clinical applicability while maintaining diagnostic performance, representing a viable technical pathway toward addressing the critical accessibility-versus-accuracy trade-off that has limited real-world implementation of AI models.

Compared to previous coding-free platforms\cite{r34,r35} and proprietary solutions like Google AutoML\cite{r28} and Microsoft Azure ML Studio\cite{r36}, GlobeReady stands out through its feature matching-based diagnostic approach, which transcends the need for fine-tuning by utilizing a comprehensive, pre-curated feature library. This design not only accelerates adaptation across diverse ophthalmic imaging modalities but also minimizes deployment and maintenance costs by eliminating extensive parameter tuning.~Furthermore, GlobeReady incorporates a unique confidence quantification system that flags low-certainty diagnostic outputs for clinician review, enhancing diagnostic reliability and safety. Its integrated case retrieval functionality facilitates rapid matching of similar cases directly from the feature library (without the need for re-training), thereby streamlining medical record tracing and supporting data annotation efforts within healthcare settings.

In addition, our prospective questionnaire-based evaluation further demonstrated that GlobeReady achieved enhanced usability, diagnostic efficiency, and clinical relevance compared to Google AutoML\cite{r28}. GlobeReady achieved significantly higher scores across all measured dimensions, including the System Usability Scale (SUS) and Likert-based assessments of effectiveness, helpfulness, ease of use, and overall satisfaction, highlighting its highly user-friendly interface tailored to the needs of clinicians. Unlike Google AutoML that require uploading patient data to cloud servers, GlobeReady is designed for seamless local deployment. This local-first architecture not only mitigates data privacy and security concerns, but also enhances accessibility, enabling clinicians to use the system without relying on complex cloud infrastructure. Notably, since GlobeReady operates as a fully training-free system, it can operate effectively in both GPU-equipped and non-GPU environments, making it particularly suitable for deployment in routine clinical settings and in regions with limited computational resources. By eliminating cloud dependencies and reducing hardware demands, GlobeReady facilitates secure, compliant, and efficient integration into real-world healthcare workflows.

Recently, several foundation models, such as UNI\cite{r37} and PLIP\cite{r38} for pathology images, RETFound\cite{r13} and VisionFM\cite{r14} for ophthalmic images, demonstrated promising performance in diverse downstream tasks. However, these models typically require fine-tuning to adapt to specific clinical contexts, posing technical barriers to widespread adoption. In contrast, GlobeReady is powered by RetiGlobe, a dedicated ophthalmic foundation model pretrained using a two-stage strategy: self-supervised learning with DINOv2\cite{r18} on 38 million synthetic ophthalmic images enriched with diverse disease patterns, followed by contrastive learning with CLIP\cite{r19} on over 475,000 globally sourced real image--text pairs. This approach enables RetiGlobe to capture fine-grained structural and pathological features while aligning visual and semantic information, resulting in robust, generalizable representations. The enhanced representational power of RetiGlobe is evidenced by t-SNE visualizations \textbf{(Fig. 3)}, where embeddings produced by GlobeReady, without any fine-tuning, show clearer inter-class separation and tighter cluster boundaries than those from RETFound\cite{r13} and VisionFM\cite{r14}.

GlobeReady's confidence-quantifiable ocular disease diagnosis further bolsters its reliability in complex clinical decision-making scenarios. Previous studies, such as UIOS\cite{r39} and FMUE\cite{r40}, explored incorporating uncertainty evaluation into disease diagnosis to enhance the reliability of AI models. However, these approaches typically require fine-tuning for specific tasks, which limits their ease of clinical adoption. Furthermore, although other studies\cite{r41,r42} have explored case-based medical image retrieval, these solutions typically function independently of diagnostic tools. In contrast, GlobeReady unifies diagnostic precision and case retrieval within a single, training-free platform. This allows for efficient clinical diagnosis across diverse ocular diseases and eliminates the need for specialized AI skills, thus offering a technical solution that fosters ease of adoption in diverse clinical settings.

Despite these promising results, several limitations warrant further study to facilitate the widespread clinical implementation of GlobeReady. Regarding data synthesis, although our dataset covers a broad spectrum of ophthalmic diseases, the inherent class imbalance, especially the underrepresentation of rare conditions, may constrain GlobeReady's performance in downstream tasks, manifesting as the relatively low Top-1 accuracy of our feature-based case retrieval function, which underscores the necessity for expanded dataset coverage. While we have augmented the dataset using a generative model to synthesize a large volume of ophthalmic images, the generation process lacked control over specific disease categories. To address this issue, a potential solution could involve exploring text-guided diffusion models\cite{r43,r44,r45} to generate category-specific samples, thereby enabling the construction of a more balanced and comprehensive dataset. Furthermore, a more rigorous evaluation of the consistency between synthetic and real-world data remains an essential research direction, particularly through formal clinician-led authenticity assessments and systematic analyses of pathological and population-specific coverage.

The DINOv2 framework was selected for self-supervised learning in this study due to its proven robustness and empirical performance across diverse vision tasks. Its contrastive teacher-student distillation mechanism is particularly effective at preserving the global structural consistency and fine pathological patterns essential for retinal imaging. While newer self-supervised paradigms continue to emerge with the potential to further enhance representation quality, this work focuses on providing a stable and validated foundation for clinician-friendly deployment.

The open set anomaly detection in our framework is designed as a crucial clinical safety function to identify cases that fall outside the reference distribution. The thresholds for this confidence-quantifiable system were determined empirically using a hold-out development subset to balance sensitivity and specificity. While the confidence-quantifiable diagnosis capability bolsters diagnostic safety, it may also flag a fraction of correctly classified, in-distribution cases for review. Under the fixed Youden-based threshold, this false-alert rate was 20\mdot{}8\% on JLHW11 (CFP) and 24\mdot{}5\% on JSIEC-OCT15 (OCT). These flags represent "needs-review" signals rather than diagnostic errors, and achieving an optimal balance between sensitivity and operational efficiency will depend on refining the threshold calibration, which can be tuned at inference time without model retraining. Although external calibration across diverse datasets including those from low-resource settings represents a potential optimization strategy, it was not performed in the current study due to data availability constraints and remains a key direction for future research.

GlobeReady represents a fundamental departure from conventional foundation models by enabling immediate training-free deployment and eliminating the need for specialized AI expertise or labor-intensive data labeling. Unlike parameter-based adaptation methods such as few-shot fine-tuning, domain adaptation, or linear probing, this non-parametric approach avoids the substantial computational resources and complex model updates that often hinder AI adoption in clinical sites with restricted infrastructure or strict regulatory constraints. Consequently, integrating these lightweight adaptation techniques into the GlobeReady framework remains a valuable direction for future research.

GlobeReady operates as a training-free system designed for high computational efficiency, enabling it to run effectively on standard clinical workstations equipped with entry-to-mid-range GPU (e.g., NVIDIA RTX 2060 or 3060 with at least 6GB-12GB VRAM). The knowledge base of the platform can be updated seamlessly by integrating newly validated cases into the reference feature library which eliminates the need for periodic and computationally expensive model retraining. This modular design enables diverse regions to expand diagnostic coverage without high-performance computing clusters thereby lowering long-term maintenance barriers in routine clinical practice. Furthermore, GlobeReady mirrors the process of case-based reasoning, and this transparency facilitates clinical application while reducing the requirement for extensive specialized training. While the current study focuses on demonstrating the technical and clinical feasibility of this approach, objective workflow-integration measurements such as time-to-diagnosis and error rates with vs. without GlobeReady were not collected; future studies should incorporate these endpoints.

Finally, the current framework primarily focuses on retinal disease diagnosis and does not support other ophthalmic tasks such as glaucoma diagnosis, nor systemic or predictive tasks such as longitudinal prognostic analysis, oculomics, or biomarker quantification (e.g., macular thickness).The core strength of GlobeReady lies in its non-parametric retrieval mechanism which in principle allows the platform to be extended to these additional tasks by populating the reference library with samples linked to comprehensive clinical phenotypes and quantitative biomarkers rather than static diagnostic labels. Notably, label quality across cohorts was safeguarded by a unified, pre-specified annotation protocol with standardized diagnostic criteria (\textbf{Appendix Tables S15-S16}), under which every annotator had to reach Cohen's kappa $\geq$ 0.8 against a senior specialist on a shared standard set before annotating local data. Because each cohort was evaluated independently, with the key comparison being within-site diagnostic performance before versus after local retrieval augmentation, a pooled cross-center inter-rater reliability statistic was not essential to the study design, and the model's stable performance across multiple independent external cohorts further supports the robustness of these labels.

In conclusion, GlobeReady is an innovative, user-friendly ophthalmic AI platform that addresses key challenges in the application of FMs in real-world clinical settings. We showed that not only does GlobeReady deliver high diagnostic accuracy and reliability across ophthalmic imaging modalities and different settings, the use and application of GlobeReady could be performed without requiring fine-tuning or AI technical expertise. GlobeReady's performance in mitigating domain shifts, coupled with high retrieval accuracy in real-world clinical evaluations, underscores its potential to improve clinician efficiency and eliminate reliance on model fine-tuning, for direct clinical application, whether it is for disease screening and monitoring and or clinical decisions support.

\subsection*{Contributors}

M.W.: conceptualization, methodology, data collection, experimental deployment, software, writing original draft. T.L.: data collection, clinical assessment and annotation and curation, software, review and editing. Q.H.: data collection and synthesis, software, review and editing. A.L.: data collection, clinical assessment and annotation and curation, review and editing. J.W. and T.C.Q,: software development. T.X.N., Q.P., D.F., C.X., Q.Y., H.Z., Z.X., G.H., H.L., T.S., M.C., Z.L., L.J.C., C.M.C., X.L., J.H., Z.X., T.F., Y.W., Q.W., W.H., Y.W., L.L., J.Y., T.H., L.L., Y.L., R.C., Y.T., X.W., Y.H., Z.Z., Q.H., J.C., M.H, C.K.M.C, A.L.Y, Z.Z., and L.L.: data collection, and performance evaluation. K.Z., T.X., Z.D.S., Y.P., L.W., Y.M., D.Z., X.W., Y.Z., C.C., D.L., J.W., X.C., C.Z., C.C.T, Y.C.T., and T.Y.W: methodology, writing---review and editing. N.T.T, Y.Z, and C.Y.C clinical support, methodology, writing---review and editing. H.F.: supervision, project administration, methodology, writing---review and editing. C-.Y.C. and H.C.: supervision, clinical assessment and annotation and curation, clinical support, writing---review and editing.

\subsection*{Acknowledgements}

This study was supported in part by the A*STAR Central Research Fund (``Robust and Trustworthy AI system for Multi-modality Healthcare''), Career Development Fund (C222812010), part by the Advanced Manufacturing and Engineering (AME) Programmatic Fund (A20H4b0141) and the National Medical Research Council, Singapore (MOH-CSASI22jul-0001), part by the Shantou Science and Technology Program (190917085269835), 2020 Li Ka Shing Foundation Cross-Disciplinary Research Grant (2020LKSFG14B), and National Key R\&D Program of China (2018YFA0701700).

\subsection*{Competing interests}

The authors declare no competing interests.

\subsection*{Code availability}

Results were further analysed and visualized with Python v.3.9, NumPy v.1.26.4, SciPy v.1.11.4, seaborn v.0.13.2, Matplotlib v.3.9.2, pandas v.2.2.2, Scikit-Learn v.1.5.1 and Pillow v.10.4.0. Heatmaps were generated with AGF-Visualization (\url{https://github.com/shirgur/AGFVisualization}). The pre-trained ophthalmic foundation model RetiGlobe is available at: \url{https://github.com/LooKing9218/RetiGlobe}. The GlobeReady platform is not planned for unrestricted public release at this stage; controlled access will be provided to qualified applicants for non-commercial research and academic use upon manuscript acceptance, subject to institutional approval and data-use agreements. To safeguard data privacy, GlobeReady is designed for local, on-site deployment within an institution's secure internal network, without transmitting patient data to external servers.

\subsection*{Data sharing}

The publicly available datasets used for pre-training are available at the following links and references:

APTOS: \url{https://www.kaggle.com/c/aptos2019-blindness-detection}.

Cataract: \url{https://www.kaggle.com/datasets/jr2ngb/cataractdataset}.

DDR: \url{https://github.com/nkicsl/DDR-dataset}.

Diabetic Retinopathy Level Detection: \url{https://www.kaggle.com/datasets/arbethi/diabetic-retinopathy-level-detection}.

Diabetic Retinopathy Organized: \url{https://www.kaggle.com/datasets/dola1507108/diabetic-retinopathy-organized}.

DR15: \url{https://www.kaggle.com/datasets/nawa393/dr15_test}.

Messidor: \url{https://paperswithcode.com/dataset/messidor-1}.

MURED: \url{https://www.kaggle.com/datasets/abhirampolisetti/multi-label-retinal-disease-mured-dataset}.

Retina Dataset: \url{https://www.kaggle.com/datasets/jr2ngb/cataractdataset}.

Kaggle DR: \url{https://www.kaggle.com/c/diabetic-retinopathy-detection/data}.

ODIR5K: \url{https://www.kaggle.com/datasets/andrewmvd/ocular-disease-recognition-odir5k}.

SIC dataset: \url{https://www.kaggle.com/competitions/innovation-challenge-2019/data}.

AliOCT2Fundus: \url{https://2024.asiateleophth.org/big-data-competition}

Retinal Rocks: \url{https://www.retinarocks.org/image-gallery}

AI-Challenge 2018: \url{https://github.com/ShawnBIT/AI-Challenger-Retinal-Edema-Segmentation}

RETOUCH dataset: \url{https://paperswithcode.com/dataset/retouch}

AliChallenge: \url{https://2024.asiateleophth.org/big-data-competition}

ACRIMA\cite{r46}, BEH\cite{r47}, DeepDRiD\cite{r48}, DR1-2\cite{r49}, E-ophta\cite{r50}, AIROGS\cite{r51}, DeepEyeNet\cite{r52}, FIVES\cite{r53}, G1020\cite{r54}, Glaucoma dataset\cite{r55,r56}, IDRiD\cite{r57}, JICHI\cite{r58}, REFUGE\cite{r59}, PARAGUAY\cite{r60}, EyePACS AirDoc\cite{r61}, JSIEC\cite{r62}, RFMid\cite{r63}, BRSET\cite{r64}, ORDRD\cite{r65}, IMDOCT\cite{r66}.

Additional data sets supporting the findings of this study are not publicly available due to the confidentiality policy of the Chinese National Health Council and institutional patient privacy regulations. However, they are available from the corresponding authors upon request. For replication of the findings and/or further academic and AI-related research activities, data may be requested from the corresponding author H.C. (haoyuchen@cuhk.edu.hk), and any requests will be responded to within 10 working days. Source data are provided in this paper.

\clearpage
\section*{References}
\renewcommand{\refname}{}

\clearpage
\begin{table}[htbp]
\centering
\caption*{\textbf{Table 1: Characteristics of ophthalmologists participating in the survey}}
\begin{tabular}{@{}p{9cm}r@{}}
\toprule
\textbf{Characteristic} & \textbf{Number (\%)} \\
\midrule
\textit{Position} & \\
\quad Intern & 5 (16.1) \\
\quad Resident without AI experience & 7 (22.6) \\
\quad Attending/Consultant without AI experience & 15 (48.4) \\
\quad Chief/Director & 4 (12.9) \\
\addlinespace
\textit{Number of images} & \\
\quad Less than 500 & 18 (58.1) \\
\quad 501-1000 & 8 (25.8) \\
\quad 1001-2000 & 1 (3.2) \\
\quad More than 2000 & 4 (12.9) \\
\addlinespace
\textit{Usage situations} & \\
\quad Daily clinical diagnosis & 19 (61.3) \\
\quad Clinical teaching & 15 (48.4) \\
\quad Labeling images & 21 (67.7) \\
\quad Other & 6 (19.4) \\
\bottomrule
\end{tabular}
\end{table}

\clearpage

\begin{table}[htbp]
\centering
\caption*{\textbf{Table 2: Questionnaire results for GlobeReady}}
\small
\begin{tabular}{@{}llccc@{}}
\toprule
\textbf{Sections} & & \textbf{\begin{tabular}[c]{@{}c@{}}Scores for GlobeReady\\ (Mean$\pm$SD)\end{tabular}} & \textbf{\begin{tabular}[c]{@{}c@{}}Scores for AutoML\\ (Mean$\pm$SD)\end{tabular}} & \textbf{\textit{P} value} \\
\midrule
\multicolumn{2}{@{}l}{\textbf{System usability scale*}} & 83.3$\pm$11.2 & 31.3$\pm$12.1 & $<$0.001 \\
Question 1 & \begin{tabular}[c]{@{}l@{}}I think that I would like to use this\\ system frequently.\end{tabular} & 3.3$\pm$0.9 & 1.1$\pm$0.9 & $<$0.001 \\
Question 2 & \begin{tabular}[c]{@{}l@{}}I found that system unnecessarily\\ complex.\end{tabular} & 3.1$\pm$1.1 & 0.9$\pm$0.8 & $<$0.001 \\
Question 3 & I thought the system was easy to use. & 3.5$\pm$0.6 & 0.9$\pm$0.6 & $<$0.001 \\
Question 4 & \begin{tabular}[c]{@{}l@{}}I think that I would need the support of a\\ technical person to be able to use this system.\end{tabular} & 3.1$\pm$1.2 & 1.0$\pm$0.7 & $<$0.001 \\
Question 5 & \begin{tabular}[c]{@{}l@{}}I found that the various functions in this\\ system were well integrated.\end{tabular} & 3.3$\pm$0.5 & 1.8$\pm$0.9 & $<$0.001 \\
Question 6 & \begin{tabular}[c]{@{}l@{}}I thought that there was too much\\ inconsistency in this system.\end{tabular} & 3.4$\pm$0.9 & 1.9$\pm$0.5 & $<$0.001 \\
Question 7 & \begin{tabular}[c]{@{}l@{}}I would imagine that most people would\\ learn to use this product very frequently.\end{tabular} & 3.4$\pm$0.6 & 1.2$\pm$0.8 & $<$0.001 \\
Question 8 & I found the system very awkward to use. & 3.6$\pm$0.8 & 1.7$\pm$0.8 & $<$0.001 \\
Question 9 & I felt very confident using the system. & 3.4$\pm$0.6 & 1.2$\pm$0.8 & $<$0.001 \\
Question 10 & \begin{tabular}[c]{@{}l@{}}I needed to learn a lot of things before I\\ could get going with this system.\end{tabular} & 3.2$\pm$0.8 & 0.8$\pm$0.6 & $<$0.001 \\
\addlinespace
\multicolumn{2}{@{}l}{\textbf{Effectiveness}} & 4.1$\pm$0.7 & 2.4$\pm$0.7 & $<$0.001 \\
Question 1 & \begin{tabular}[c]{@{}l@{}}I spend less time in clinical diagnosis by\\ utilizing the system.\end{tabular} & 4.1$\pm$0.9 & 2.1$\pm$0.9 & $<$0.001 \\
Question 2 & \begin{tabular}[c]{@{}l@{}}The results generated by the system are\\ trustworthy and accurate.\end{tabular} & 4.1$\pm$0.8 & 2.7$\pm$0.8 & $<$0.001 \\
Question 3 & \begin{tabular}[c]{@{}l@{}}It would cost me more time to read and\\ correct the diagnosis results.\end{tabular} & 4.0$\pm$0.9 & 2.4$\pm$0.9 & $<$0.001 \\
\addlinespace
\multicolumn{2}{@{}l}{\textbf{Helpfulness}} & & & \\
Question & The system can assist me in clinical diagnosis. & 4.4$\pm$0.9 & 2.5$\pm$1.1 & $<$0.001 \\
\addlinespace
\multicolumn{2}{@{}l}{\textbf{Ease of use}} & 4.4$\pm$0.6 & 2.5$\pm$0.6 & $<$0.001 \\
Question 1 & \begin{tabular}[c]{@{}l@{}}The user interface is clear and easy to\\ understand.\end{tabular} & 4.4$\pm$0.6 & 2.5$\pm$0.8 & $<$0.001 \\
Question 2 & \begin{tabular}[c]{@{}l@{}}I find it difficult to use the system as its\\ guidance is unclear.\end{tabular} & 4.4$\pm$0.9 & 2.5$\pm$1.1 & $<$0.001 \\
\addlinespace
\multicolumn{2}{@{}l}{\textbf{Satisfaction}} & 4.5$\pm$0.7 & 2.1$\pm$0.7 & $<$0.001 \\
Question 1 & \begin{tabular}[c]{@{}l@{}}I would like to recommend my colleagues\\ to use this system.\end{tabular} & 4.5$\pm$0.7 & 2.1$\pm$0.8 & $<$0.001 \\
Question 2 & I would use this system in my future work. & 4.5$\pm$0.7 & 2.2$\pm$0.7 & $<$0.001 \\
\bottomrule
\multicolumn{5}{@{}p{15.5cm}@{}}{\footnotesize *System usability scale score was calculated by multiplying the score for each question (Questions 1 to 10) by 2.5, and the max score for each question is 4.} \\
\end{tabular}
\end{table}

\clearpage
\begin{figure}[p]
  \centering
  \includegraphics[width=\textwidth]{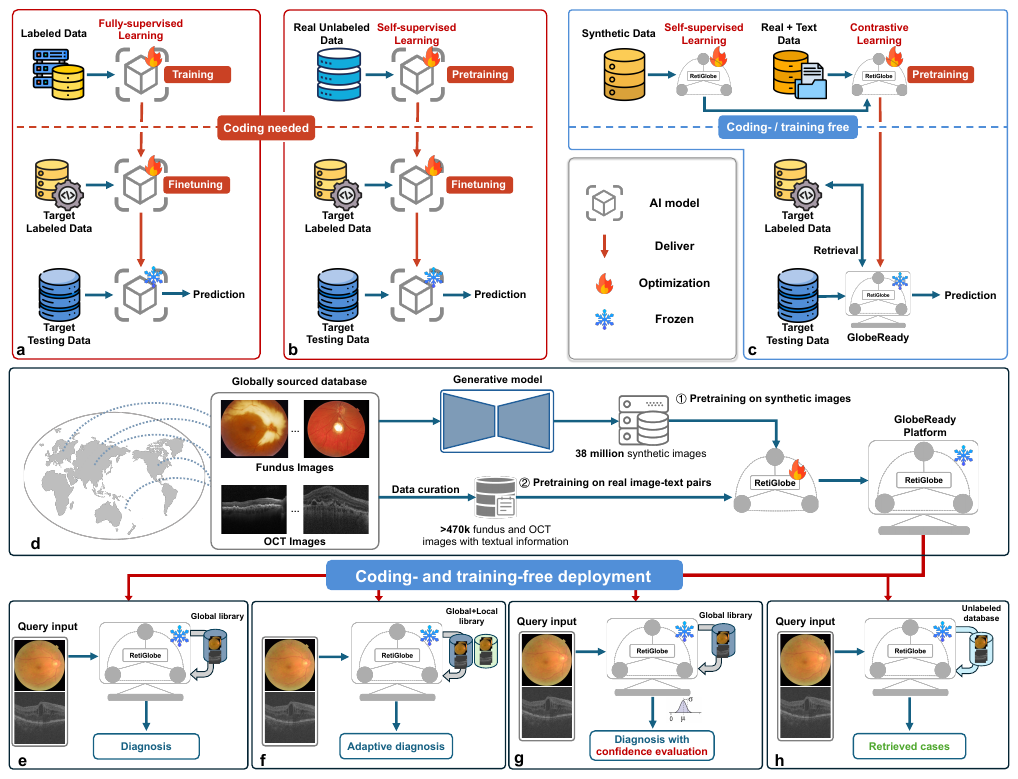}
  \caption*{\textbf{Fig. 1 | Overview of the framework. a,} Traditional deep learning approaches: Full-supervised training on labeled real data and finetune to target tasks; \textbf{b,} Foundation model: Self-supervised pre-training on massive real unlabeled data and finetune to target tasks; \textbf{c}, Self-supervised pretraining on a combination of synthetic and real image+text, followed by direct deployment to target tasks without the need for fine-tuning; \textbf{d,} Data collection and model pre-training. \textbf{e,} GlobeReady disease diagnosis; \textbf{f,} Adaptive ocular disease diagnosis with local retrieval augmentation; \textbf{g,} Training-free confidence-quantifiable ocular disease diagnosis; \textbf{h,} Feature-based case retrieval.}
  \label{fig:1}
\end{figure}

\begin{landscape}
\begin{figure}[p]
  \centering
  \includegraphics[width=0.98\linewidth]{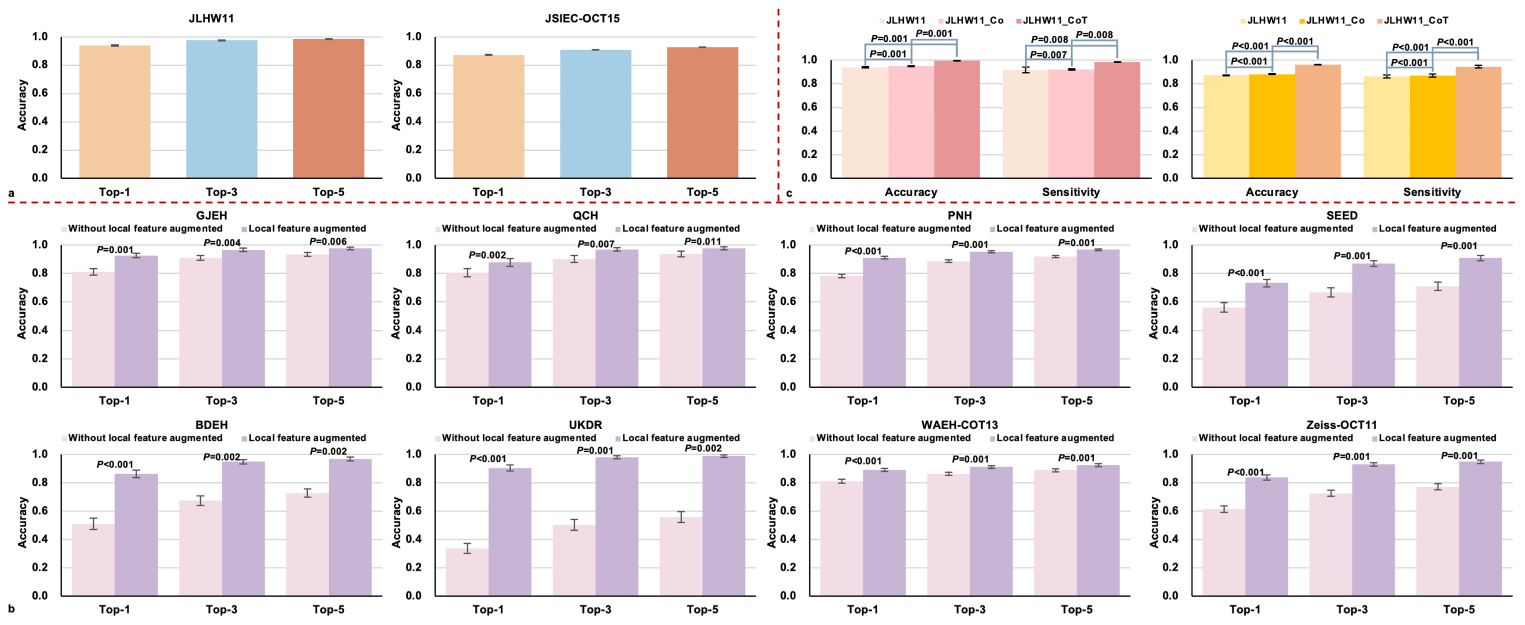}
  \caption*{\textbf{Fig. 2 | a. Performance of GlobeReady for identifying ocular diseases on different datasets. }JLHW11 is a CFP dataset from the \textbf{J}oint Shantou International Eye Center (JSIEC), \textbf{L}ongchuan People's Hospital, \textbf{H}aifeng Pengpai Memorial Hospital, and \textbf{W}uhan Aier Eye Hospital (WAEH), covering 11 categories. JSIEC-OCT15 is an OCT dataset created by Joint Shantou International Eye Center with 15 categories. \textbf{b. Results of local retrieval augmented ocular disease diagnosis. }The Guangxi Jingliang Eye Hospital (\textbf{GJEH}) dataset includes the same disease categories as JLHW11; \textbf{QCH} dataset, from Qingdao Central Hospital, comprises images labeled as normal, AMD, DR, and glaucoma. PNH dataset, from Puning People's Hospital, also shares the same categories as JLHW11. The Singapore Epidemiology of Eye Diseases (\textbf{SEED}) dataset, collected in Singapore, includes annotations for diagnosing AMD, DR, and glaucoma. \textbf{BDEH} dataset originates from Binh Dinh Eye Hospital in Vietnam. \textbf{UKDR} dataset, was collected from a DR screening program in the University of Liverpool, the United Kingdom. \textbf{WAEH-OCT13} dataset is a local OCT dataset from the Wuhan Aier Eye Hospital. The \textbf{Zeiss-OCT11} dataset is a local OCT dataset acquired by the device of Zeiss.\textbf{ c. Confidence-quantifiable ocular disease diagnosis. }JLHW11 and JSIEC-OCT15 represent the GlobeReady's performance without introducing confidence-quantification; JLHW11\_Co and JSIEC-OCT15\_Co indicate the GlobeReady's performance with confidence-quantification; JLHW11\_CoT and JSIEC-OCT15\_CoT are the performance of GlobeReady after implementing a confidence threshold recommendation. In this last scenario, samples exhibiting low confidence are flagged for re-evaluation by a clinician.}
  \label{fig:2}
\end{figure}
\end{landscape}

\begin{figure}[p]
  \centering
  \includegraphics[width=\textwidth]{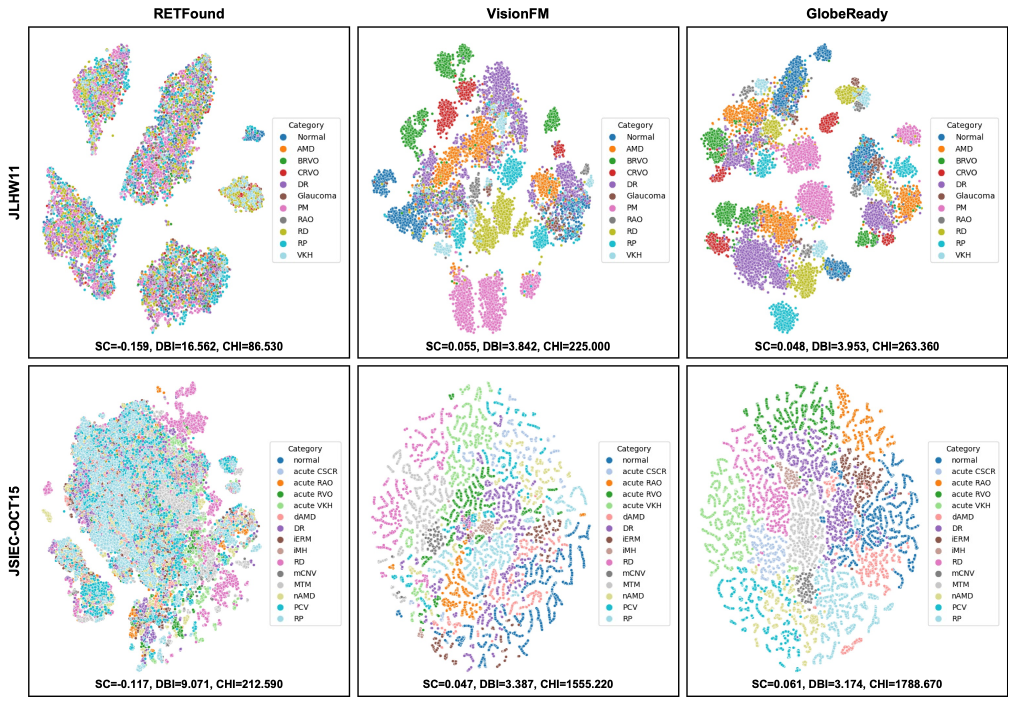}
  \caption*{\textbf{Fig. 3 | t-SNE visualization of feature embeddings extracted by different pretrained foundation models.} Points with different colors represent samples from different disease categories.}
  \label{fig:3}
\end{figure}

\begin{figure}[p]
  \centering
  \includegraphics[width=0.82\textwidth]{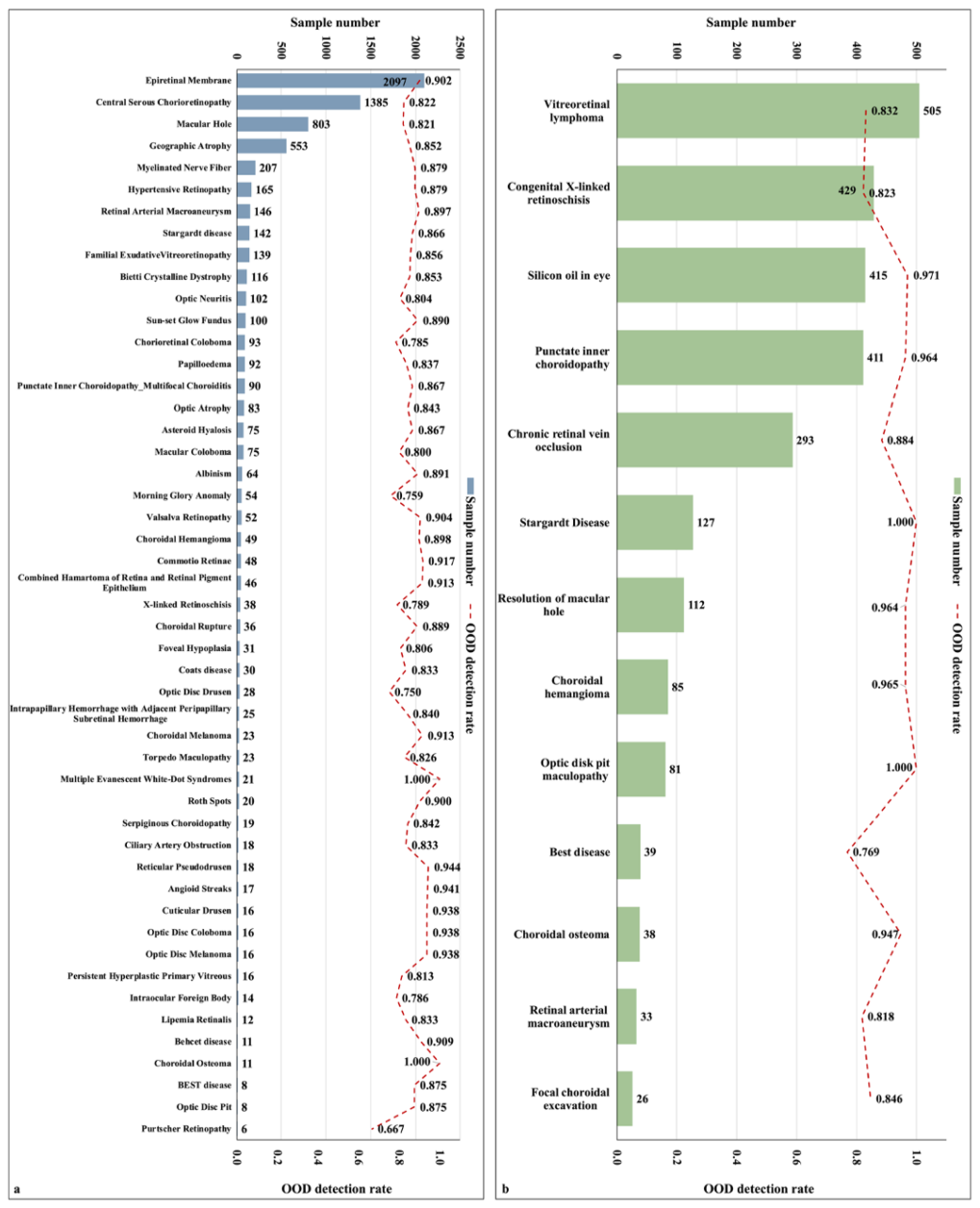}
  \caption*{\textbf{Fig. 4 | GlobeReady's performance in detecting out-of-distribution (OOD) diseases in an open-set scenario. a,} Data distribution and corresponding OOD detection rates for the 49 disease categories in CFPOOD49 dataset. \textbf{b,} Data distribution and corresponding OOD detection rates for the 13 disease categories in OCT-OOD13 dataset.}
  \label{fig:4}
\end{figure}

\begin{figure}[p]
  \centering
  \includegraphics[width=\textwidth]{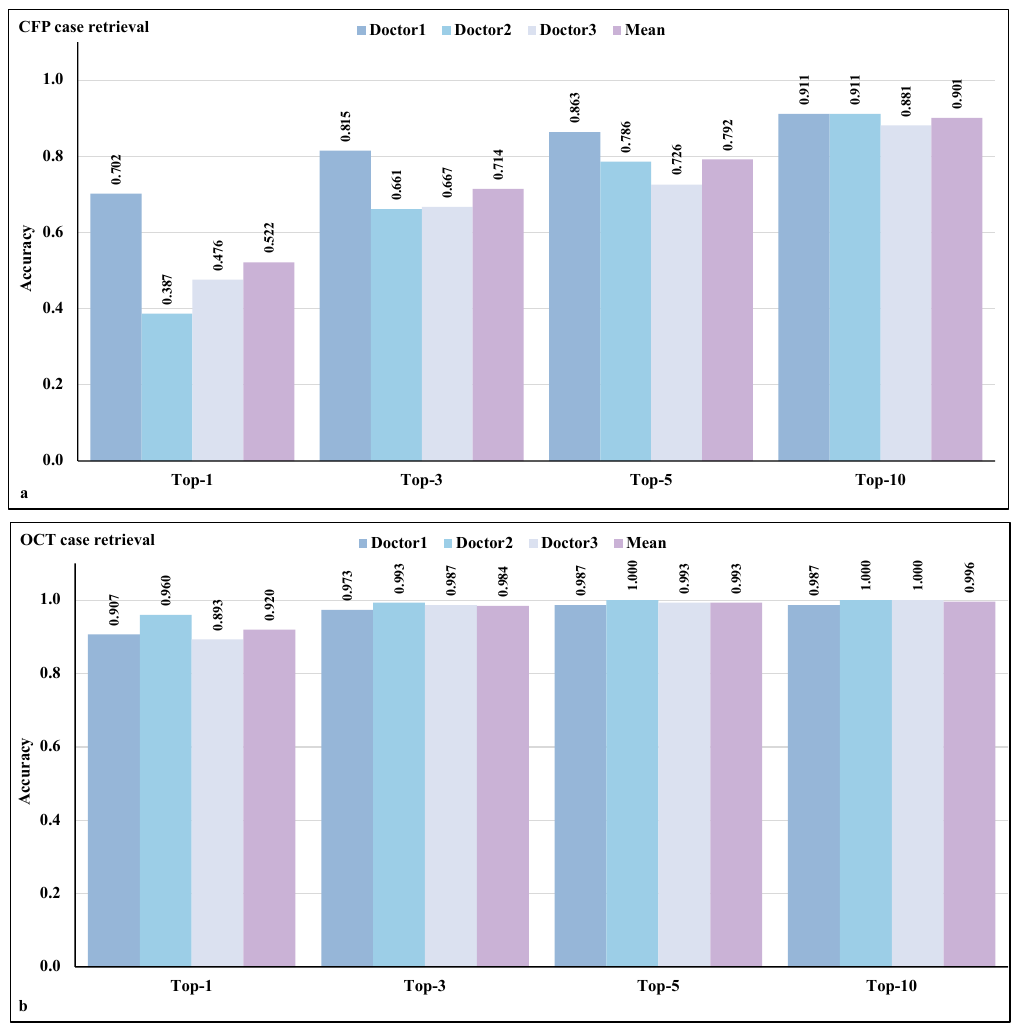}
  \caption*{\textbf{Fig. 5 | Top-k score in prospective evaluation of feature-based case retrieval. }a, The prospective evaluation results of feature-based case retrieval for CFPs; b, The prospective evaluation results of feature-based case retrieval for OCT images. Doctor1, Doctor2, and Doctor3 refer to the three ophthalmologists who participated in the prospective evaluation.}
  \label{fig:5}
\end{figure}

\begin{figure}[p]
  \centering
  \includegraphics[width=\textwidth]{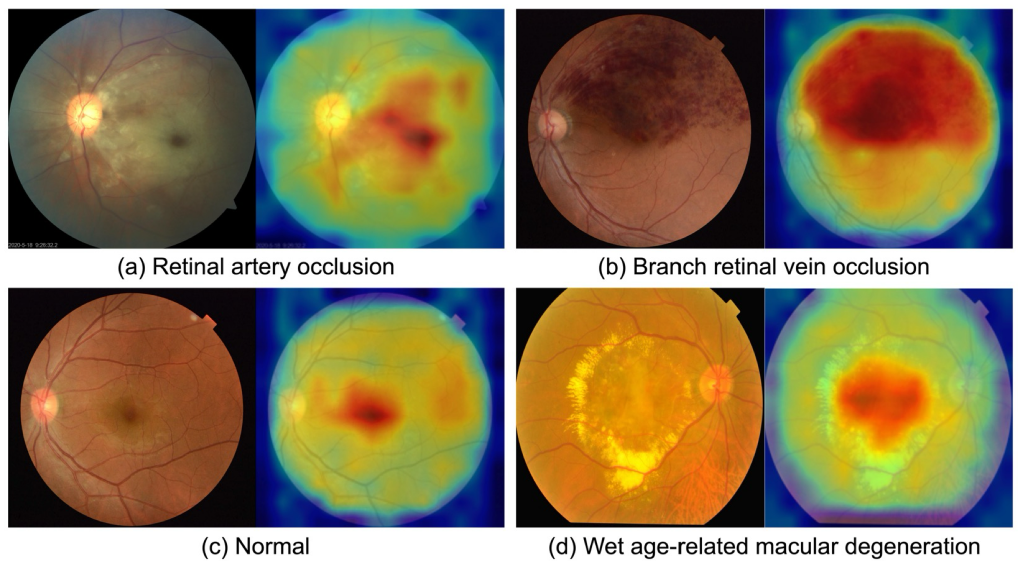}
  \caption*{\textbf{Fig. 6 | Heatmaps for various diseases in CFPs. }These heatmaps show that the model automatically focuses on clinically relevant pathological lesions across conditions in typical cases, including (a) retinal artery occlusion, (b) branch retinal vein occlusion, (c) normal cases, and (d) wet age-related macular degeneration.}
  \label{fig:6}
\end{figure}

\begin{figure}[p]
  \centering
  \includegraphics[width=\textwidth]{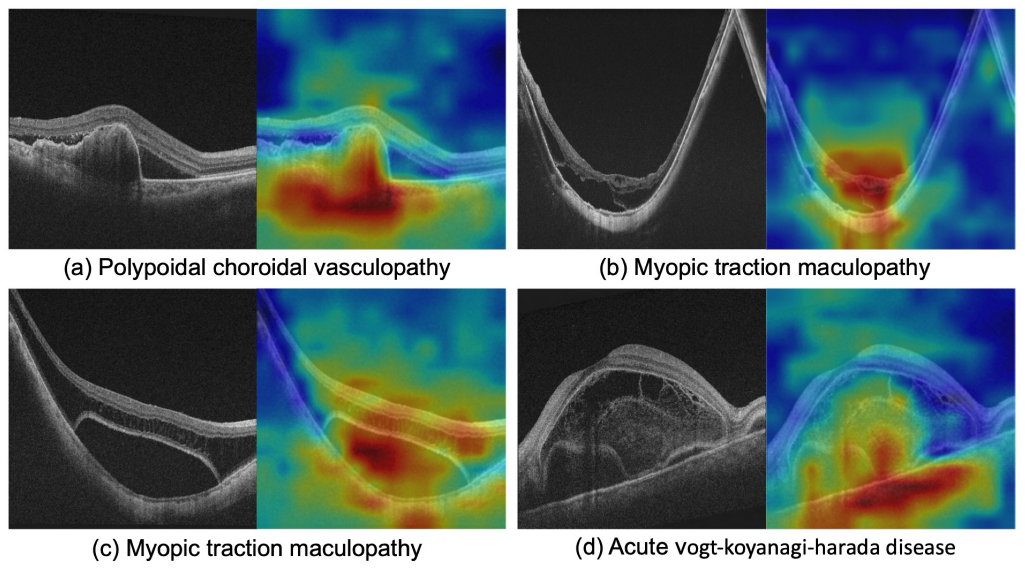}
  \caption*{\textbf{Fig. 7 | Heatmaps for various diseases in OCT images. }These heatmaps show that the model intrinsically localizes specific structural abnormalities across conditions in typical cases, including (a) polypoidal choroidal vasculopathy, (b-c) myopic traction maculopathy, and (d) acute Vogt-Koyanagi-Harada disease.}
  \label{fig:7}
\end{figure}


\begin{thebibliography}{99}
\setlength{\itemsep}{2pt}
\bibitem{r1} Nusinovici S, Rim T H, Li H, et al. Application of a deep-learning marker for morbidity and mortality prediction derived from retinal photographs: a cohort development and validation study. The Lancet Healthy Longevity, 2024, 5(10).
\bibitem{r2} Zhang K, Zhou R, Adhikarla E, et al. A generalist vision--language foundation model for diverse biomedical tasks. Nature Medicine, 2024: 1-13.
\bibitem{r3} Dai L, Wu L, Li H, et al. A deep learning system for detecting diabetic retinopathy across the disease spectrum. Nature communications, 2021, 12(1): 3242.
\bibitem{r4} Peng Y, Zhu W, Chen Z, et al. Automatic staging for retinopathy of prematurity with deep feature fusion and ordinal classification strategy. IEEE transactions on medical imaging, 2021, 40(7): 1750-1762.
\bibitem{r5} Hosny A, Parmar C, Quackenbush J, et al. Artificial intelligence in radiology. Nature Reviews Cancer, 2018, 18(8): 500-510.
\bibitem{r6} F\"orsch S, Klauschen F, Hufnagl P, et al. Artificial intelligence in pathology. Deutsches \"Arzteblatt International, 2021, 118(12): 199.
\bibitem{r7} Li Z, Wang L, Wu X, et al. Artificial intelligence in ophthalmology: The path to the real-world clinic. Cell Reports Medicine, 2023, 4(7).
\bibitem{r8} Dai L, Sheng B, Chen T, et al. A deep learning system for predicting time to progression of diabetic retinopathy. Nature Medicine, 2024, 30(2): 5.
\bibitem{r9} Yan Q, Weeks D E, Xin H, et al. Deep-learning-based prediction of late age-related macular degeneration progression. Nature machine intelligence, 2020, 2(2): 141-150.
\bibitem{r10} Burton M J, Ramke J, Marques A P, et al. The lancet global health commission on global eye health: vision beyond 2020. The Lancet Global Health, 2021, 9(4): e489-e551.
\bibitem{r11} Song G, Jelly E T, Chu K K, et al. A review of low-cost and portable optical coherence tomography. Progress in Biomedical Engineering, 2021, 3(3): 032002.
\bibitem{r12} Kumar V, Paul K. Fundus imaging-based healthcare: Present and future. ACM Transactions on Computing for Healthcare, 2023, 4(3): 1-34.
\bibitem{r13} Zhou Y, Chia M A, Wagner S K, et al. A foundation model for generalizable disease detection from retinal images. Nature, 2023, 622(7981): 156-163.
\bibitem{r14} Qiu J, Wu J, Wei H, et al. Visionfm: a multi-modal multi-task vision foundation model for generalist ophthalmic artificial intelligence. NEJM AI 2024;1(12).
\bibitem{r15} Esteva A, Robicquet A, Ramsundar B, et al. A guide to deep learning in healthcare. Nature medicine, 2019, 25(1): 24-29.
\bibitem{r16} Bommasani R, Hudson D A, Adeli E, et al. On the opportunities and risks of foundation models. arXiv preprint arXiv:2108.07258, 2021.
\bibitem{r17} Wiggins W F, Tejani A S. On the opportunities and risks of foundation models for natural language processing in radiology. Radiology: Artificial Intelligence, 2022, 4(4): e220119.
\bibitem{r18} Oquab M, Darcet T, Moutakanni T, et al. Dinov2: Learning robust visual features without supervision. arXiv preprint arXiv:2304.07193, 2023.
\bibitem{r19} Radford A, Kim J W, Hallacy C, et al. Learning transferable visual models from natural language supervision. International conference on machine learning. PmLR, 2021: 8748-8763.
\bibitem{r20} Hou Q, Wang M, Cao P, Zou K, Liu X, Fu H, Zaiane OR. FundusGAN: A Hierarchical Feature-Aware Generative Framework for High-Fidelity Fundus Image Generation..
\bibitem{r21} Wang M, Lin T, Lin A, et al. Enhancing diagnostic accuracy in rare and common fundus diseases with a knowledge-rich vision-language model. Nature Communications, 2025, 16(1): 5528.
\bibitem{r22} Dosovitskiy A, Beyer L, Kolesnikov A, et al. An image is worth 16x16 words: Transformers for image recognition at scale. arXiv preprint arXiv:2010.11929, 2020.
\bibitem{r23} Hu E J, Shen Y, Wallis P, et al. Lora: Low-rank adaptation of large language models. ICLR, 2022, 1(2): 3.
\bibitem{r24} Majithia S, Tham Y C, Chee M L, et al. Cohort profile: the Singapore epidemiology of eye diseases study (seed). International journal of epidemiology, 2021, 50(1): 41-52.
\bibitem{r25} Gal Y, Ghahramani Z. Dropout as a bayesian approximation: Representing model uncertainty in deep learning. International conference on machine learning. PMLR, 2016: 1050-1059.
\bibitem{r26} Kendall A, Gal Y. What uncertainties do we need in bayesian deep learning for computer vision?. Advances in neural information processing systems, 2017, 30.
\bibitem{r27} Luo J, Xiong C. Youden index and associated cut-points for three ordinal diagnostic groups. Communications in Statistics-Simulation and Computation, 2013, 42(6): 1213-1234.
\bibitem{r28} Bisong E, Bisong E. Google automl: cloud vision. Building Machine Learning and Deep Learning Models on Google Cloud Platform: A Comprehensive Guide for Beginners, 2019: 581-598.
\bibitem{r29} Brooke J. SUS: A "quick and dirty" usability scale. In: Patrick W. Jordan BT, Bernard A. Weerdmeester, Ian L. McClelland, ed. Usability evaluation in industry. Taylor \& Francis; 1996:189-194.
\bibitem{r30} Bangor A, Kortum PT, Miller JT. Determining what individual SUS scores mean: adding an adjective rating scale. Journal of Usability Studies archive. 2009;4:114-123.
\bibitem{r31} Lewis P, Perez E, Piktus A, et al. Retrieval-augmented generation for knowledge-intensive nlp tasks. Advances in neural information processing systems, 2020, 33: 9459-9474.
\bibitem{r32} Ma X, Gong Y, He P, et al. Query rewriting in retrieval-augmented large language models. Proceedings of the 2023 Conference on Empirical Methods in Natural Language Processing. 2023: 5303-5315.
\bibitem{r33} Gur S, Ali A, Wolf L. Visualization of supervised and self-supervised neural networks via attribution guided factorization. Proceedings of the AAAI conference on artificial intelligence. 2021, 35(13): 11545-11554
\bibitem{r34} Milad D, Antaki F, Mikhail D, et al. Code-Free Deep Learning Glaucoma Detection On Color Fundus Images. Ophthalmology Science, 2025: 100721.
\bibitem{r35} Wagner S K, Liefers B, Radia M, et al. Development and international validation of custom-engineered and code-free deep-learning models for detection of plus disease in retinopathy of prematurity: a retrospective study. The Lancet Digital Health, 2023, 5(6): e340-e349.
\bibitem{r36} Barnes J. Azure machine learning. Microsoft Azure Essentials. 1st ed, Microsoft, 2015.
\bibitem{r37} Chen R J, Ding T, Lu M Y, et al. Towards a general-purpose foundation model for computational pathology. Nature Medicine, 2024, 30(3): 850-862.
\bibitem{r38} Huang Z, Bianchi F, Yuksekgonul M, et al. A visual--language foundation model for pathology image analysis using medical twitter. Nature medicine, 2023, 29(9): 2307-2316.
\bibitem{r39} Wang M, Lin T, Wang L, et al. Uncertainty-inspired open set learning for retinal anomaly identification. Nature Communications, 2023, 14(1): 6757.
\bibitem{r40} Peng Y, Lin A, Wang M, et al. Enhancing AI reliability: A foundation model with uncertainty estimation for optical coherence tomography-based retinal disease diagnosis. Cell Reports Medicine, 2025, 6(1).
\bibitem{r41} Fang, J., Fu, H., Liu, J. Deep triplet hashing network for case-based medical image retrieval. Medical image analysis, 69, 101981 (2021).
\bibitem{r42} Quellec, G., Lamard, M., Cazuguel, G., Bekri, L., Daccache, W., et al. Automated assessment of diabetic retinopathy severity using content-based image retrieval in multimodal fundus photographs. Investigative ophthalmology \& visual science, 52, 8342-8348 (2011).
\bibitem{r43} Ramesh A, Pavlov M, Goh G, et al. Zero-shot text-to-image generation. International conference on machine learning. Pmlr, 2021: 8821-8831.
\bibitem{r44} Ramesh A, Dhariwal P, Nichol A, et al. Hierarchical text-conditional image generation with clip latents. arXiv preprint arXiv:2204.06125, 2022, 1(2): 3.
\bibitem{r45} Betker J, Goh G, Jing L, et al. Improving image generation with better captions. Computer Science. \url{https://cdn.openai.com/papers/dall-e-3.pdf}, 2023, 2(3): 8.
\bibitem{r46} Diaz-Pinto, A., Morales, S., Naranjo, V., et al. CNNs for automatic glaucoma assessment using fundus images: an extensive validation. Biomedical engineering online, \textbf{18,} 1-19 (2019).
\bibitem{r47} Islam, M. T., Mashfu, S. T., Faisal, A., Siam, S. C., Naheen, I. T., et al. Deep learning-based glaucoma detection with cropped optic cup and disc and blood vessel segmentation. \textit{Ieee Access}, \textbf{10,} 2828-2841 (2021).
\bibitem{r48} Liu R, Wang X, Wu Q, Dai, L., Fang, X., Yan, T., et al. Deepdrid: Diabetic retinopathy---grading and image quality estimation challenge. \textit{Patterns}, \textbf{3, }(2022).
\bibitem{r49} Pires, R., Jelinek, H. F., Wainer, J., Valle, E., Rocha, A. Advancing bag-of-visual-words representations for lesion classification in retinal images. \textit{PloS one}, \textbf{9,} e96814 (2014).
\bibitem{r50} Decenciere, E., Cazuguel, G., Zhang, X., Thibault, G., Klein,J.-C., et al. TeleOphta: Machine learning and image processing methods for teleophthalmology. \textit{Irbm}, \textbf{34,} 196-203 (2013).
\bibitem{r51} De Vente, C., Vermeer, K. A., Jaccard, N., Wang, H., Sun, H., et al. AIROGS: artificial intelligence for robust glaucoma screening challenge. IEEE transactions on medical imaging, \textbf{43,} 542-557 (2023).
\bibitem{r52} Huang, J. H., Yang, C. H. H., Liu, F., Tian, M., Liu, Y. -C., et al. Deepopht: medical report generation for retinal images via deep models and visual explanation. \textit{Proceedings of the IEEE/CVF winter conference on applications of computer vision}, 2442-2452 (2021).
\bibitem{r53} Jin, K., Huang, X., Zhou, J., Li, Y., Yan, Y., et al. Fives: A fundus image dataset for artificial Intelligence based vessel segmentation. \textit{Scientific data}, \textbf{9,} (2022).
\bibitem{r54} Bajwa, M. N., Singh, G. A. P., Neumeier, W., Malik, M. I., Dengel, A., et al. G1020: A benchmark retinal fundus image dataset for computer-aided glaucoma detection. \textit{2020 International Joint Conference on Neural Networks (IJCNN 2020).} 1-7 (2020).
\bibitem{r55} Singh, A., Dutta, M. K., ParthaSarathi, M., Uher, V., Burget, R. Image processing based automatic diagnosis of glaucoma using wavelet features of segmented optic disc from fundus image. Computer methods and programs in biomedicine, \textbf{124,}108-120 (2016).
\bibitem{r56} Kermany D S, Goldbaum M, Cai W, et al. Identifying medical diagnoses and treatable diseases by image-based deep learning. cell, 2018, 172(5): 1122-1131. e9.
\bibitem{r57} Porwal P, Pachade S, Kokare M, et al. Idrid: Diabetic retinopathy--segmentation and grading challenge. Medical image analysis, 2020, 59: 101561.
\bibitem{r58} Takahashi H, Tampo H, Arai Y, et al. Applying artificial intelligence to disease staging: Deep learning for improved staging of diabetic retinopathy. PloS one, 2017, 12(6): e0179790.
\bibitem{r59} Orlando J I, Fu H, Breda J B, et al. Refuge challenge: A unified framework for evaluating automated methods for glaucoma assessment from fundus photographs. Medical image analysis, 2020, 59: 101570.
\bibitem{r60} Ben\'\i{}tez V E C, Matto I C, Rom\'an J C M, et al. Dataset from fundus images for the study of diabetic retinopathy. Data in brief, 2021, 36: 107068.
\bibitem{r61} Ju L, Wang X, Wang L, et al. Improving medical images classification with label noise using dual-uncertainty estimation. IEEE transactions on medical imaging, 2022, 41(6): 1533-1546.
\bibitem{r62} Cen, L. P., Ji, J., Lin, J. W., Ju, S., Lin, H., et al. Automatic detection of 39 fundus diseases and conditions in retinal photographs using deep neural networks. \textit{Nature communications,} \textbf{12,} 4828 (2021).
\bibitem{r63} Pachade S, Porwal P, Thulkar D, et al. Retinal fundus multi-disease image dataset (rfmid): A dataset for multi-disease detection research. Data, 2021, 6(2): 14.
\bibitem{r64} Nakayama LF, Restrepo D, Matos J, Ribeiro LZ, Malerbi FK, Celi LA, Regatieri CS. BRSET: A Brazilian Multilabel Ophthalmological Dataset of Retina Fundus Photos. PLOS Digit Health. 2024 Jul 11;3(7):e0000454. doi: 10.1371/journal.pdig.0000454. PMID: 38991014; PMCID: PMC11239107.
\bibitem{r65} Adal K.M., van Etten P.G., Martinez J.P., van Vliet L.J., Vermeer K.A. Accuracy Assessment of Intra- and Intervisit Fundus Image Registration for Diabetic Retinopathy Screening. Investig. Ophthalmol. Vis. Sci. 2015;56:1805--1812.
\bibitem{r66} Kermany D S, Goldbaum M, Cai W, et al. Identifying medical diagnoses and treatable diseases by image-based deep learning. cell, 2018, 172(5): 1122-1131. e9.
\bibitem{r67} Kulyabin, M., Zhdanov, A., Nikiforova, A. et al. OCTDL: Optical Coherence Tomography Dataset for Image-Based Deep Learning Methods. Scientific data, 11(1): 365.
\bibitem{r68} Gholami, P., Roy, P., Parthasarathy, M. K., \& Lakshminarayanan, V. OCTID: Optical coherence tomography image database. Computers \& Electrical Engineering, 2020, 81: 106532.
\bibitem{r69} Sim\'eoni O, Vo H V, Seitzer M, et al. Dinov3: arXiv preprint arXiv:2508.10104, 2025.
\end{thebibliography}
\end{document}